\documentclass[journal]{IEEEtran}

\title{Towards Real-Time Monocular Depth Estimation for Robotics: A Survey}

\author{Xingshuai Dong, Matthew A. Garratt,~\IEEEmembership{Senior Member,~IEEE}, Sreenatha G. Anavatti, and Hussein A. Abbass,~\IEEEmembership{Fellow,~IEEE} 
	
School of Engineering and Information Technology, University of New South Wales, Canberra, Australia.
}

\usepackage{amsmath}
\usepackage{amssymb}
\usepackage{cite}
\usepackage{color}
\usepackage{comment}
\usepackage{epsfig}
\usepackage{float}
\usepackage{colortbl}
\usepackage{graphicx}
\usepackage{geometry}
\usepackage[pagewise]{lineno}
\usepackage{makecell}
\usepackage{mathrsfs}
\usepackage{ragged2e}
\usepackage{subfig}

\graphicspath{{Images/}}
\DeclareGraphicsExtensions{.pdf,.png,.jpg,.eps}

\begin{document}
	
\maketitle

\title{Lineno in the Abstract}
\begin{abstract}
As an essential component for many autonomous driving and robotic activities such as ego-motion estimation, obstacle avoidance and scene understanding, monocular depth estimation (MDE) has attracted great attention from the computer vision and robotics communities. Over the past decades, a large number of methods have been developed. To the best of our knowledge, however, there is not a comprehensive survey of MDE. This paper aims to bridge this gap by reviewing 197 relevant articles published between 1970 and 2021. In particular, we provide a comprehensive survey of MDE covering various methods, introduce the popular performance evaluation metrics and summarize publically available datasets. We also summarize available open-source implementations of some representative methods and compare their performances. Furthermore, we review the application of MDE in some important robotic tasks. Finally, we conclude this paper by presenting some promising directions for future research. This survey is expected to assist readers to navigate this research field.
\end{abstract}


\IEEEpeerreviewmaketitle
	
\section{Introduction}
\IEEEPARstart{D}{epth} estimation refers to the process of estimating a dense depth map from the corresponding input image(s). Depth information can be utilized to infer the 3D structure, which is an essential part in many robotics and autonomous system tasks, such as ego-motion estimation \cite{tateno2017cnn}, obstacle avoidance \cite{yang2019fast} and scene understanding \cite{jiang2018self}. Active methods depend on RGB-D cameras, LIDAR, Radar or ultrasound devices to directly get the depth information of the scene \cite{forouher2016sensor}. However, RGB-D cameras suffer from a limited measurement range, LIDAR and Radar is limited to sparse coverage, and ultrasound devices are limited by the inherently imprecise measurements. In addition, the above devices are large in size and energy-consuming, which is a deficiency when it comes to small sized robots such as micro aerial vehicles (MAVs).  \par
On the contrary, RGB cameras are cheaper and light weight. More importantly, they can provide richer information about the environment. Many methods \cite{wang2011global, muresan2015improving, spangenberg2014large} depend on stereo matching to estimate depth maps from stereo images. Stereo methods are more accurate, however, collecting stereo images require complex alignment and calibration procedures. Besides, stereo vision-based methods are limited by the baseline distance between the two cameras. To be specific, the estimated depth values tend to be inaccurate when the considered distance are large. With the advancements in computer vision algorithms, it is more convenient to infer a dense depth map from RGB images. \par
\begin{figure}[h]
	\centering
	\includegraphics[width=.9\linewidth]{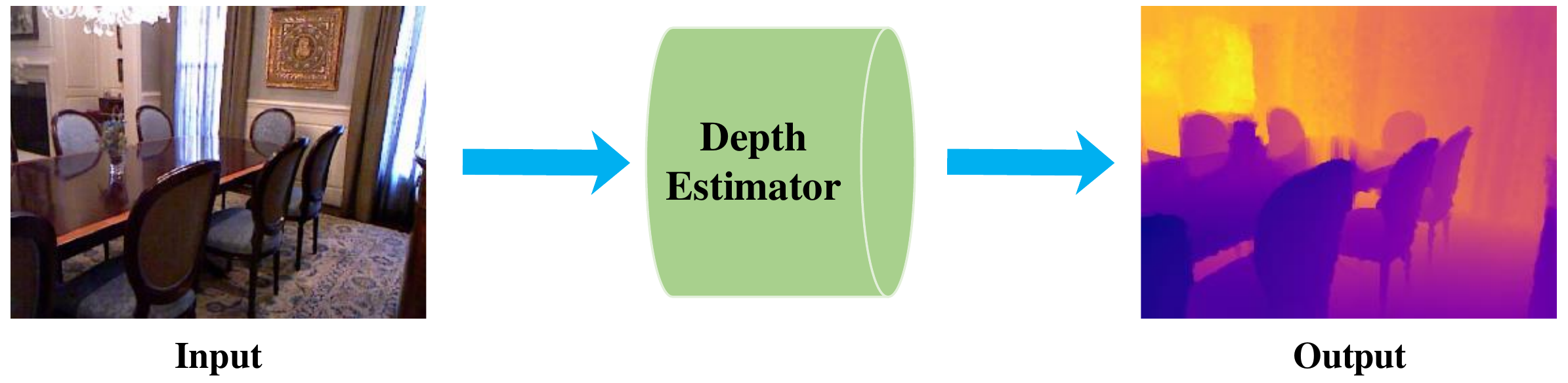} \\
	\caption{The overall pipeline of monocular depth estimation. Given an input RGB image, the depth predictor produces a dense depth map (best viewed in color). (The colors in the depth map correspond to the distance of that pixel, yellow and blue indicates far and close respectively.)}
	\label{fig:overallpipeline}
\end{figure}
\begin{figure}[tb]
	\centering
	\includegraphics[width=.9\linewidth]{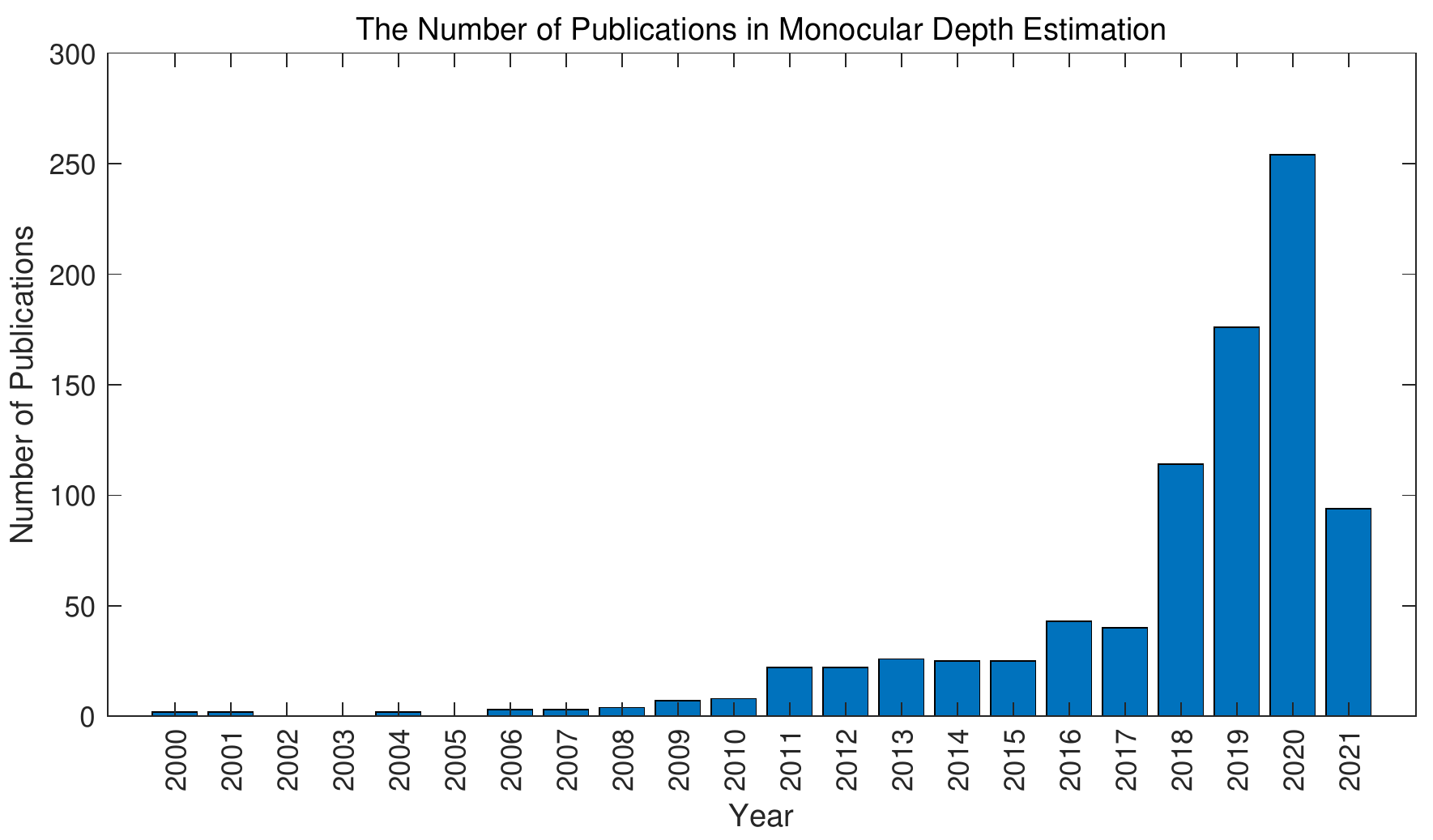} \\
	\caption{The number of published articles on monocular depth estimation from 2000 to June 2021. (Data from Google scholar advanced search.)}
	\label{fig:Number_of_Published_Papers_in_MDE}
\end{figure}
\begin{figure*}[tb]
	\centering
	\includegraphics[width=.9\linewidth]{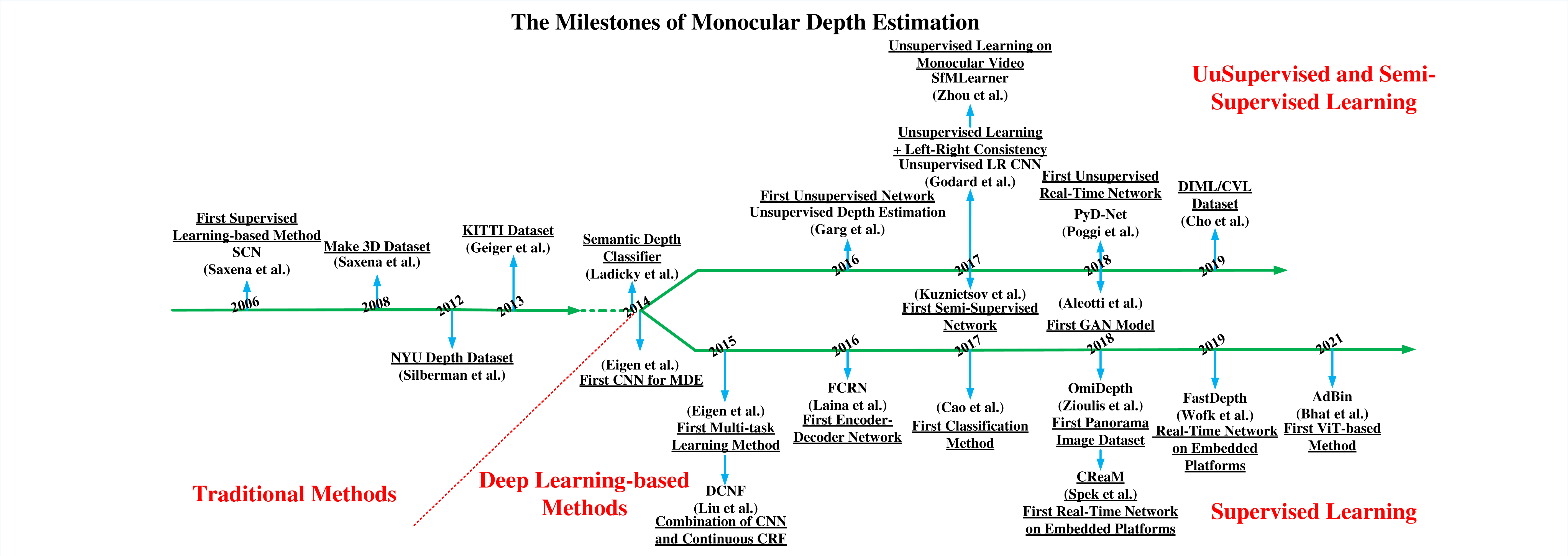} \\
	\caption{Milestones of monocular depth estimation, including traditional handcrafted feature based-methods \cite{saxena2006learning, saxena2008make3d, silberman2012indoor, geiger2013vision, ladicky2014pulling} and state-of-the-art deep learning-based methods \cite{eigen2014depth, eigen2015predicting, liu2015learning, laina2016deeper, garg2016unsupervised, godard2017unsupervised, zhou2017unsupervised, kuznietsov2017semi, aleotti2018generative, poggi2018towards, zioulis2018omnidepth, spek2018cream, cho2019large, wofk2019fastdepth, bhat2021adabins}.}
	\label{fig:milestones}
\end{figure*}

In this paper, we restrict our survey to monocular depth estimation (MDE) for dense depth maps. We extensively survey 197 relevant articles spanning over 50 years (from 1970 to 2021). Our aim is to assist readers to navigate this research field, which has attracted great attention from the computer vision and robotics communities. Figure \ref{fig:Number_of_Published_Papers_in_MDE} shows the number of published articles on MDE from 2000 to 2021, while Figure \ref{fig:milestones} illustrates the milestones of MDE in recent years. \par
We classify the reviewed methods into three main categories: Structure from Motion (SfM)-based methods, traditional handcrafted feature-based methods, and state-of-the-art deep learning-based methods. SfM-based methods \cite{wedel2006realtime, prakash2014sparse, ha2016high, javidnia2017accurate} track a set of corresponding pixels, across a series of images taken in a given scene, and compute depth values at the pixels where features are matched. Therefore, the obtained depth maps are sparse. For traditional handcrafted feature-based methods \cite{torralba2002depth, saxena2006learning, jung2010depth, liu2010single, ladicky2014pulling, liu2014discrete, karsch2014depth, raza2015depth}, features are first extracted from monocular images, which are then utilized to estimate dense depth maps by optimizing a probabilistic model such as a Markov Random Field (MRF) or a Conditional Random Field (CRF). Over the past few years, the success of deep neural networks (DNNs) has greatly motivated the development of MDE. A variety of models \cite{eigen2014depth, eigen2015predicting, laina2016deeper, cao2017estimating, xu2017multi, li2018monocular, hu2019revisiting, wofk2019fastdepth} manifest their effectiveness to recover the depth information from a single image. A possible reason is that the monocular cues can be better modeled with the larger capacity of DNNs. \par
\begin{figure}[tb]
	\centering
	\includegraphics[width=.9\linewidth]{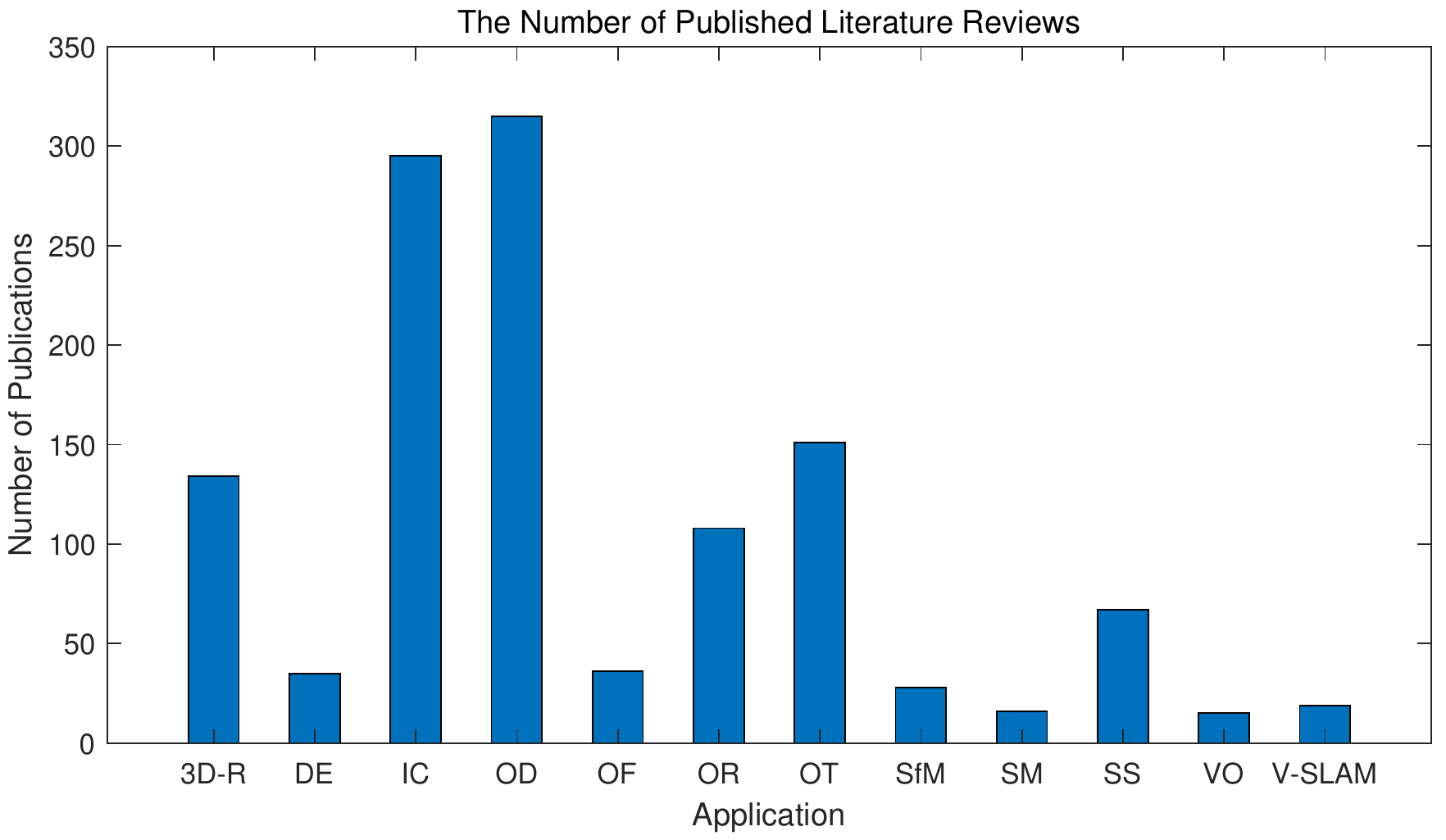} \\
	
	\caption{The number of published literature reviews on different applications until June 2021. (``3D-R": 3D Reconstruction, ``DE'': Depth Estimation, ``IC": Image Classification, ``OD": Object Detection, ``OF": Optical Flow Estimation, ``OT": Object Tracking, ``SfM": Structure from Motion, ``SM": Stereo Matching, ``SS": Semantic Segmentation, ``VO": Visual Odometry and ``V-SLAM": Visual Simultaneous Localization and Mapping.) (Data from Google scholar advanced search.)} 
	\label{fig:Number of Published Literature Reviews}
\end{figure}

In recent years, a number of surveys concerning depth estimation \cite{chavan2015depth, mohan2015review, jamwal2016survey, bahadur2017literature, bhoi2019monocular, laga2019survey, vyshna2019literature, liu2020survey} have been published, as summarized in Table \ref{table:summary_of_reviews}. According to Figure \ref{fig:Number of Published Literature Reviews}, compared with applications such as 3D reconstruction, SfM, stereo matching, etc, fewer surveys on depth estimation have been published. However, among these existing surveys on depth estimation, a comprehensive one covering classic and deep learning-based methods as well as the application of MDE is lacking. Therefore, this survey aims to bridge this gap with the following contributions: \par
\begin{itemize}
	\item {\textbf{Focusing the survey entirely on MDE:}} Most of the previous surveys \cite{chavan2015depth, mohan2015review, jamwal2016survey, bahadur2017literature, laga2019survey, vyshna2019literature, liu2020survey} blend monocular and stereo algorithms together, and hence do not provide a comprehensive survey of MDE. Moreover, \cite{bhoi2019monocular} only reviews 5 state-of-the-art deep learning-based methods and does not reveal the technical evolution of MDE. This paper performs an extensive investigation into the methods of MDE from the SfM and traditional handcrafted feature-based methods to the state-of-the-art deep learning-based methods. In addition, we summarize the publically available datasets, commonly used performance metrics and open-source implementations of the representative methods. \par  
	\item {\textbf{Conduct a comprehensive survey in the aspect of MDE for robotics:}} Previous surveys only focus on the techniques of depth estimation. As a main method for range perception, MDE plays an important role in the field of robotics. This paper makes an investigation to the application of MDE for ego-motion estimation, obstacle avoidance and scene understanding. We illustrate how MDE works on the above tasks and make robots more intelligent. \par
\end{itemize}
\par
\begin{figure*}[tb]
	\centering
	\includegraphics[width=.9\linewidth]{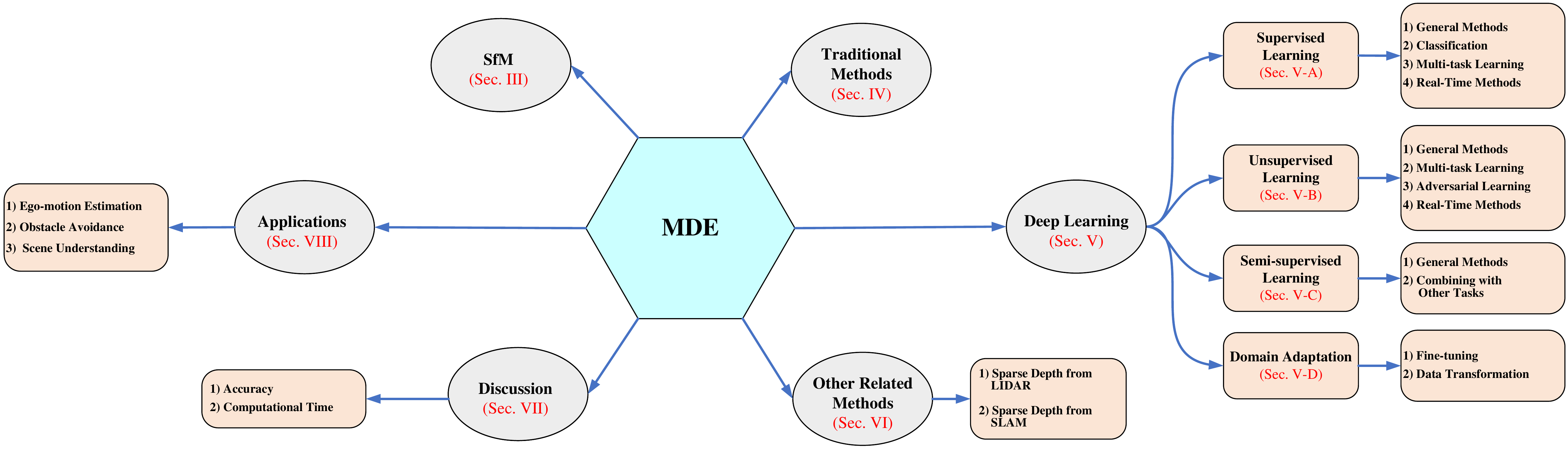} \\
	\caption{An overview of the organization of this paper (from Section III to Section VIII.). (``MDE": Monocular Depth Estimation and ``SfM": Structure from Motion.)}
	\label{fig:overallorganization}
\end{figure*}

The remainder of this paper is organized as follows: In Section 2, we introduce the background of MDE. MDE with Structure from Motion and traditional handcrafted feature-based methods will be reviewed in Section 3 and Section 4 respectively. Section 5 reviews state-of-the-art deep learning-based methods. Section 6 will review other related methods. Section 7 presents a discussion and comparison on different MDE methods. The applications of monocular depth estimation will be reviewed in Section 8. Conclusions are given in Section 9. We show the overall organization of this paper in Fig. \ref{fig:overallorganization}. \par
\begin{table*}[h]
	\renewcommand{\arraystretch}{1.3}	
	\caption{Summarization of a number of related surveys on depth estimation in recent years. ``Ref.'': Reference.}
	\centering
	\begin{tabular}{c | c | c | c | c | c } \hline \hline
		\textbf{No.} & \textbf{Title} & \textbf{Ref.} & \textbf{Year} & \textbf{Published} & \textbf{Description} \\ \hline 
		1 & Depth Extraction from Video: A Survey & \cite{chavan2015depth} & 2015 & IJIRAE & \makecell{A survey on depth estimation from videos and propose a \\ method for generating depth maps from videos and images} \\ \hline
		2 & \makecell {A Review on Depth Estimation for \\ Computer Vision Applications} & \cite{mohan2015review} & 2015 & Position & \makecell{A survey of different methods for depth estimation} \\ \hline
		3 & A Survey on Depth Map Estimation Strategies & \cite{jamwal2016survey} & 2016 & ICSP & \makecell{A survey of active and passive methods for depth estimation} \\ \hline
		4 & \makecell{Literature Review on Various Depth \\ Estimation Methods for An Image} & \cite{bahadur2017literature} & 2017 & IJRG & \makecell{A survey on different depth estimation \\ methods using cues from two images} \\ \hline
		5 & \makecell{Monocular Depth Estimation: A Survey}  & \cite{bhoi2019monocular} & 2019 & arXiv & \makecell{A review of five algorithms of MDE \\ with different deep learning techniques} \\ \hline 
		6 & \makecell{A Survey on Deep Learning Architectures \\ for Image-based Depth Reconstruction} & \cite{laga2019survey} & 2019 & arXiv& \makecell{A comprehensive survey of recent developments \\ in depth reconstruction with deep learning techniques } \\ \hline
		7 & \makecell{Literature Review on Depth Estimation \\ Using a Single Image} & \cite{vyshna2019literature} & 2019 & IJARIIT & \makecell{A review of learning-based MDE} \\ \hline
		8 & \makecell{A Survey of Depth Estimation \\ Based on Computer Vision} & \cite{liu2020survey} & 2020 & DSC & \makecell{A survey of current mainstream active \\ and passive depth approaches} \\ \hline  
		9 & \textbf{\makecell{This Survey}} & - & \textbf{2021} & \textbf{Ours} & \textbf{\makecell{A comprehensive survey of MDE from classic to deep \\ learning-based methods, and its application in robotics}} \\ \hline \hline 
	\end{tabular} \\
	\justify ``DSC": International Conference on Data Science in Cyberspace, ``ICSP": International Conference on Signal Processing, ``IJRG": International Journal of Research-Granthaalayah, ``IJIRAE": International Journal of Innovative Research in Advanced Engineering and ``IJARIIT": International Journal of Advance Research, Ideas and Innovations in Technology.
	\label{table:summary_of_reviews}
\end{table*}
\section{Background of Monocular Depth Estimation}
\subsection{Problem Definition}
The problem definition of monocular depth estimation (MDE) can be viewed as follows. Let $I$ be a single RGB image with size $w \times h$, $D$ is the corresponding depth map with the same size as $I$. The task of MDE is to formulate a non-linear mapping $\Psi$: $I \to D$. Whilst requiring less computational resources and avoiding the baseline issue, MDE is an ill-posed problem as a monocular image may be captured from different distinct 3D scenes. Therefore, MDE algorithms exploit different monocular cues such as texture, occlusion, known object size, lighting, shading, haze and defocus. \par
\subsection{Performance Evaluation}
Given an estimated depth map $D$ and the corresponding ground-truth $D^*$, let $D_i$ and $D^*_i$ represent the estimated and ground-truth depth values at the pixel indexed by $i$, respectively, and $N$ represent the total number of pixels for which there exist both valid ground-truth and estimated depth pixels. As for the quantitative comparison of the estimated depth map and ground-truth, the commonly used evaluation metrics in prior works are listed as follows:
\begin{itemize}
	\item {\textbf{Absolute Relative Difference (Abs Rel)}}: defined as the average value over all the image pixels of the $L_1$ distance between the ground-truth and the estimated depth, but scaled by the estimated depth:
	\begin{equation} \label{eq:abs_rel}
	Abs Rel = \frac{1}{N}\sum_{N}\frac{|{D^*_i} - {D_i}|}{{D_i}} 
	\end{equation}
	\item {\textbf{Squared Relative Difference (Sq Rel)}}: defined as the average value over all the image pixels of the $L_2$ distance between the ground-truth and the estimated depth, but scaled by the estimated depth:
	\begin{equation} \label{eq:sq_rel}
	Sq Rel = \frac{1}{N}\sum_{N}\frac{|{D^*_i} - {D_i}|^2}{{D_i}} 
	\end{equation}
	\item {\textbf{The linear Root Mean Square Error (RMSE)}}: defined as:
	\begin{equation} \label{eq:rmse}
	RMSE = \sqrt{\frac{1}{N}\sum_{N}|{D^*_i} - {D_i}|^2} 
	\end{equation}
	\item {\textbf{The logarithm Root Mean Square Error (RMSE log)}}: defined as:
	\begin{equation} \label{eq:rmse_log}
	RMSE\ log = \sqrt{\frac{1}{N}\sum_{N}|log\ {D^*_i} - log\ {D_i}|^2} 
	\end{equation}
	\item {\textbf{Accuracy under a threshold}:} is the percentage of predicted pixels where the relative error is within a threshold. The formula is represented as:
	\begin{equation} \label{eq:threshold}
	max(\frac{D_i}{D^*_i}, \frac{D^*_i}{D_i}) < threshold
	\end{equation}
	where the values of threshold usually set to 1.25, $1.25^2$, $1.25^3$.
\end{itemize} \par
In addition, Eigen et al. \cite{eigen2014depth} design a scale-invariant error to measure the relationships between points in the scene, irrespective of the absolute global scale. The scale-invariant mean squared error in log space is defined in Equation \ref{eq:scale_invariant_error}:
\begin{equation} \label{eq:scale_invariant_error}
E(D, {D^*}) = \frac{1}{2N} \sum_{i=1}^{N}({log{\,}{D}_{i}} - {log {\,}{D^*}_{i}} + \alpha (D, {D^*}))^2
\end{equation}
where $\alpha(D, {D^*}) = \frac{1}{N} \sum_{i}({log{\,}{D^*}_{i}} - {log{\,}{D}_{i}})$ is the value of $\alpha$ which minimizes the difference for a given $(D, {D^*})$. For any estimation $D$, $err^{\alpha}$ is the scale that best aligns it to the ground-truth,  and $err$ is the difference between $D$ and $D^{*}$ in log space. Since all scalar multiples of ${D^{*}}$ have the same error, the scale is invariant. \par
\subsection{Datasets}
Datasets play a critical role in developing and evaluating depth estimation. In depth estimation, a number of well-known datasets have been released. Beginning with Make3D \cite{saxena2008make3d}, representative datasets include NYU depth v2 \cite{silberman2012indoor}, KITTI \cite{geiger2013vision}, Cityscapes \cite{cordts2016cityscapes} and Virtual KITTI \cite{gaidon2016virtual, cabon2020virtual}. The features of the datasets for depth estimation are summarized in Table \ref{table: datasets_for_mde}.
\begin{table*}[tb]
	\renewcommand{\arraystretch}{1.3}
	\caption{A summary of the datasets for depth estimation. ``K": thousand, ``M": million and ``-": not available.}
	\centering
	\begin{tabular}{c | c | c | c | c | c | c | c} \hline\hline
		Year & Dataset & Scenario & Sensors & Resolution & Type & Images & Annotation \\ \hline\hline
		2008 & Make3D \cite{saxena2008make3d} & Outdoor & Laser Scanner & $2272\times1704$ & Real & 534 & Dense \\ \hline
		2012 & NYU-v2 \cite{silberman2012indoor} & Indoor & Kinect v1 & $640\times480$ & Real & 1449 & Dense \\ \hline
		2012 & RGB-D SLAM \cite{sturm2012benchmark} & Indoor & Kinect v1 & $640\times480$ & Real & 48K & Dense \\ \hline 
		2013 & KITTI \cite{geiger2013vision} & Driving & LiDAR & $1238\times374$ & Real & 44K & Sparse \\ \hline  
		2015 & SUN RGB-D \cite{song2015sun} & Indoor & - & - & Real & 10335 & Dense \\ \hline
		2016 & DIW \cite{chen2016single} & Outdoor & - & - & Real & 495K & Single Pair \\ \hline
		2016 & Cityscapes \cite{cordts2016cityscapes} & Driving & Stereo Camera & $2048\times1024$ & Real & 5000 & Disparity \\ \hline 
		2016 & CoRBS \cite{wasenmuller2016corbs} & Indoor & Kinect v2 & \makecell{$1920\times1080$ for RGB, \\ $512\times424$ for Depth} & Real & - & Dense \\ \hline
		2016 & Virtual KITTI \cite{gaidon2016virtual} & Outdoor & - & $1242\times375$ & Synthetic & 21260 & Dense \\ \hline
		2017 & 2D-3D-S \cite{armeni2017joint} & Indoor & Matterport Camera & $1080\times1080$ & Real & 71909 & Dense \\ \hline
		2017 & ETH3D \cite{schops2017multi} & In/Outdoor & Laser Scanner & $940\times490$ & Real & - & Dense \\ \hline
		2017 & Matterport3D \cite{chang2017matterport3d} & Indoor & Matterport Camera & $1280\times1024$ & Real & 194400 & Dense \\ \hline
		2017 & ScanNet \cite{dai2017scannet} & Indoor & Structure Sensor & \makecell{$1296\times968$ for RGB, \\ $640\times480$ for Depth} & Real & 2.5M & Dense \\ \hline
		2017 & SceneNet RGB-D \cite{mccormac2017scenenet} & Indoor & - & $320\times240$ & Synthetic & 5M & Dense \\ \hline
		2017 & SUNCG \cite{song2017semantic} & Indoor & - & $640\times480$ & Synthetic & 45000 & Dense \\ \hline
		2018 & MegaDepth \cite{li2018megadepth} & In/Outdoor & - & - & Real & 130K & Dense/Ordinal \\ \hline
		2018 & Unreal \cite{mancini2018j} & Outdoor & - & $256\times160$ & Synthetic & 107K & Dense \\ \hline
		2018 & SafeUAV \cite{marcu2018safeuav} & Outdoor & - & $640\times480$ & Synthetic & 8137 & Dense \\ \hline
		2018 & 3D60 \cite{zioulis2018omnidepth} & Indoor & - & - & Synthetic & 35995 & Dense \\ \hline
		2018 & NUSTMS \cite{wu2019depth} & Outdoor & Radar & \makecell{$576\times160$ for Infrared, \\ $144\times40$ for Depth} & Real & 3600 & Dense \\ \hline
		2019 & DIML/CVL \cite{cho2019large} & In/Outdoor & \makecell{Kinect v2, \\ Zed Stereo Camera} & \makecell{$1920\times1080$ \\ $1280\times720$} & Real & 1M & Dense \\ \hline 
		2019 & DrivingStereo \cite{yang2019drivingstereo} & Driving & LiDAR & $1762\times800$ & Real & 182K & Sparse \\ \hline
		2019 & DIODE \cite{vasiljevic2019diode} & In/Outdoor & Laser Scanner & $ 1024\times768 $ & Real & 25458 & Dense \\ \hline
		2019 & Mid-Air \cite{fonder2019mid} & Outdoor & - & $1024\times1024$ & Synthetic & 119K & Dense \\ \hline
		2020 & \makecell{Forest Environment} \cite{niu2020low} & Forest & Depth Camera & $640\times480$ & Real & 134K & Dense \\ \hline
		2020 & Shanghaitech-Kujiale \cite{jin2020geometric} & Indoor & - & $1024\times512$ & Synthetic & 3500 & Dense \\ \hline
		2020 & Virtual KITTI 2 \cite{cabon2020virtual} & Outdoor & - & $1242\times375$ & Synthetic & 21260 & Dense \\  
		\hline \hline
	\end{tabular}
	\label{table: datasets_for_mde}	
\end{table*}
\section{Structure from Motion Based Methods}
Structure from motion (SfM) refers to the process of predicting camera motion and/or 3D structure of the environment from a sequence of images taken from different viewpoints \cite{ullman1979interpretation}. Given a sequence of input images that taken from different viewpoints, features such as Harris, SIFT, or SURF are first extracted from all the images. Then the extracted features will be matched. Because some features maybe incorrectly matched, RANSAC (random sample consensus) is typically applied to remove the outliers. These matched features are tracked from image to image to estimate the 3-D coordinates of the features. This produces a point cloud which can be transformed to a depth map. \par
\subsection{Methodologies}
Wedel et al. \cite{wedel2006realtime} estimate the scene depth from the scaling of supervised image regions using SfM. Define a point ${\textsl{\textbf{X}(t)} = (X(t), Y(t), Z(t))^{T}}$ in 3-D space at time $t$, and its corresponding projected image point $\textsl{\textbf{x}(t)}$. The camera translation in depth between time $t$ and ${t + \tau}$ is $\textsl{\textbf{T}(t, $\tau$)}$, while the point at time ${t + \tau}$ is  $\textsl{\textbf{X}(t + $\tau$)} = \textsl{\textbf{X}(t)} + \textsl{\textbf{T}(t, $\tau$)}$. The motion of image regions are divided into two parts, with the correctly computed vehicle translation and displacement of image points, the scene depth can be computed through Equation \ref{eq:d_f}:
\begin{equation} \label{eq:d_f}
d\equiv{Z(t)} = \frac{s(t, \tau)}{1 - s(t, \tau)}T_{Z}(t, \tau)   
\end{equation}
where $s(t, \tau)$ is the scale, and $T_{Z}(t, \tau)$ is the camera translation in depth. \par
Prakash et al. \cite{prakash2014sparse} present a SfM-based sparse depth estimation method. The proposed approach takes a sequence of 5 to 8 images captured by a monocular camera to estimate a depth map. With the captured images, features are detected by a multi-scale Fast detector. After matching the detected features from a reference frame and any other frame in the input subset, the two-view geometry is computed between the considered frames. The sparse depth values at the matched feature locations are calculated and reconstructed through a metric transformation. \par 
Ha et al. \cite{ha2016high} propose a Structure from Small Motion (SfSM) method which utilizes a plane sweep technique to estimate a depth map. The Harris corner detector is applied to extract features in the reference image and the corresponding features in other images are found by the Kanade-Lukas-Tomashi (KLT) algorithm. Then, the plane sweeping technique is applied to get dense depth maps. Although \cite{ha2016high} generates dense depth maps, it takes about 10 minutes to process just one image. To solve the problem of computing efficiency, Javidnia and Corcoran \cite{javidnia2017accurate} utilize the ORB algorithm as the feature extractor. It reduces the run time to minutes but still cannot run in real-time. In addition, as a corner detector, the ORB algorithm is highly sensitive to the texture present in the scene. As a result, the estimated depth maps are erroneous in low-textured environments. \par 
\subsection{Summary}
SfM relies on feature detection and matching to get the correspondence between the detected features and the accuracy of produced depth values depends on the quality of feature matching. The number of detected features relies on the environment, for example, less features are detected in textureless or low-contrast surroundings. Therefore, most of the existing SfM methods produce the sparse depth maps. These depth maps are adequate for the task of localization, but are not sufficient for applications such as autonomous flight which requires a dense depth map to enable UAVs to avoid frontal obstacles. Although using more features and images produces better estimations, it requires more time to generate a depth map. \par
\section{Traditional Handcrafted Feature Based Methods}
Due to the loss of 3D information in the process of capturing images with a monocular camera, it is not straightforward to infer a depth map from a single-view image. Unlike stereo vision-based methods that can perform stereo matching between the left and right images to estimate depth, earlier monocular depth estimation (MDE) algorithms mainly use texture variations, occlusion boundaries, defocus, color/haze, surface layout and size of known objects as cues for predicting depth maps. Although Markov Random Field (MRF) and its variants are a branch of machine learning, they are often combined with handcrafted features to incorporate more contextual information. Therefore, we review methods with MRF in this section to distinguish from the deep neural network (DNN)-based methods. \par
The handcrafted feature based-methods roughly work as follows. First, the input images are over-segmented into a set of small regions, called superpixels. Each such superpixel is assumed as a coherent region in the scene that all the pixels have similar properties. Then a number of color, location, texture, motion and geometric context-based features are computed from the obtained superpixels. With the computed features, depth cues will be computed to estimate the depth for each superpixel. Finally, a MRF model is applied to combine superpixel-based depth estimation with information between different superpixels to construct the final depth map. \par
\subsection{Methodologies}
According to Google scholar, the pioneering work in depth estimation from monocular images is \cite{horn1970shape}. In this work, the intensity or color gradients of a monocular image are exploited to estimate the depth information of objects. The intrinsic images correspond to physical properties of the scene such as depth, reflectance, shadows and surface shape, provide complementary information \cite{barrow1978recovering}. Inspired by this point, Kong and Black \cite{kong2015intrinsic} formulate dense depth estimation as an intrinsic image estimation problem. They combine \cite{karsch2014depth} with a method that extracts consistent albedo and shading from monocular video. A contour detector is trained to predict surface boundaries from albedo, shading and pixel values and the predicted contour is applied to replace image boundaries to enhance the qualities of depth maps. \par 
Torralba and Oliva \cite{torralba2002depth} propose the first learning-based approach, which infers absolute depth from monocular images by incorporating the size of known objects in the image. As the recognition of objects under unconstrained conditions is difficult and unreliable, the absolute scene depth of the images is derived from the global image structure represented as a set of features from Fourier and wavelet transforms. Real-world images contain various objects, while the work in \cite{torralba2002depth} handles different objects with the same method. Hence, it is unsuitable because it disregards the object's own properties. Jung and Ho \cite{jung2010depth} design an MDE algorithm using a Bayesian learning-based object classification method. With the property of linear perspective, objects in a monocular image are categorized into four types: sky, ground, cubic and plane. According to the type, a relative depth value to each object and 3D model is generated. \par  
Saxena et al. \cite{saxena2006learning} introduce a supervised learning-based method to estimate depth from monocular images. They divide the input image into small patches and estimate a single depth value for each patch. Two kinds of features, absolute and relative depth features are applied. The former is used to estimate the absolute depth at a particular patch and the latter is for distinguishing the depth magnitude between two patches. Considering the depth of a particular patch relies on the features of the patch and the depths of other parts of the image, a MRF is utilized to model the relation between the depth of a patch and the depths of its neighbouring patches. Raza et al. \cite{raza2015depth} combine the texture features, geometric context, motion boundary-based monocular cues with co-planarity, connectivity and spatio-temporal consistency constraints to infer depth from monocular videos. Given a monocular video, they first decompose it into spatio-temporal regions. For each region, depth cues that model the relationship of depth to visual appearance, motion and geometric classes are computed and utilized to estimate depth with random forest regression. Subsequently, the estimated depth is refined by incorporating 3D scene properties in MRF with occlusion boundaries. \par 
Besides image features, semantic labels are also used as a cue for inferring depth. The semantic classes of a pixel or region usually have geometry constraints, for example, sky is far away and ground is horizontal. Therefore, depth can be estimated by measuring the difference in appearance with respect to a given semantic class. Liu et al. \cite{liu2010single} propose a method that uses semantic information as context to estimate depth from a single image. The proposed method consists of two steps. In the first step, a learned multi-class image labeling MRF is applied to infer the semantic class for each pixel in the image. The obtained semantic information is incorporated in the depth reconstruction model in the second step. Two different MRF models, a pixel-based and a superpixel-based, are designed. Both MRF models define convex objectives that are solved by using the L-BFGS algorithm to compute a depth value for each pixel in the image. \par
Ladicky et al. \cite{ladicky2014pulling} demonstrate how semantic labeling and depth estimation can benefit each other under a unified framework. They propose a pixel-wise classifier by using the property of perspective geometry. Conditioning the semantic label on the depth promotes the learning of a more discriminative classifier. Conditioning depth on semantic classes enables the classifier to overcome some ambiguities of depth estimation. The relationship between different parts of the image is another cue for estimating depth. Liu et al. \cite{liu2014discrete} model MDE as a discrete-continuous optimizing problem. The continuous variables encode the depth of the superpixels in the input image, and the additional discrete variables encode the relationship of two neighboring superpixels. With these variables, the depth estimation can be solved by an inference problem in a discrete-continuous CRF. \par
Karsch et al. \cite{karsch2014depth} design a non-parametric, data-driven method for estimating depth maps from 2D videos or single images. Given a new image, the designed algorithm first searches similar images from a dataset by applying GIST matching. Subsequently, the label transfer between the given image and the matched image are applied to construct a set of possible depth values for the scene. Finally, the spatio-temporal regularization in an MRF formulation is conducted to make the generated depths spatially smooth. \par 
\subsection{Summary}
In the above mentioned methods, handcrafted features are extracted from the monocular images to estimate depth maps by optimizing a probabilistic model. These features are designed beforehand by human experts to extract a given set of chosen characteristics, while some corner cases may be missed. Therefore, it may result in unsatisfactory performance when applied in new environments. In addition, these methods need pre-processing or post-processing, which imposes a computational burden and makes them unsuitable for the real-time control of robots. \par
\section{Deep Learning Based Methods}
The success of deep learning in image classification also boosts the development of monocular depth estimation (MDE). In this section, we review deep learning based-MDE methods. According to the dependency on ground-truth, there are three types of learning approaches: supervised, unsupervised and semi-supervised. These three types of methods are trained on the real data, we also review methods trained on the synthetic data and then transferred to the real data in the fourth sub-section. The implementations and download links of the source code of some algorithms are summarized in Table \ref{table: implementations_of_mde}. \par
\begin{figure*}[tb]
	\centering
	\includegraphics[width=.9\linewidth]{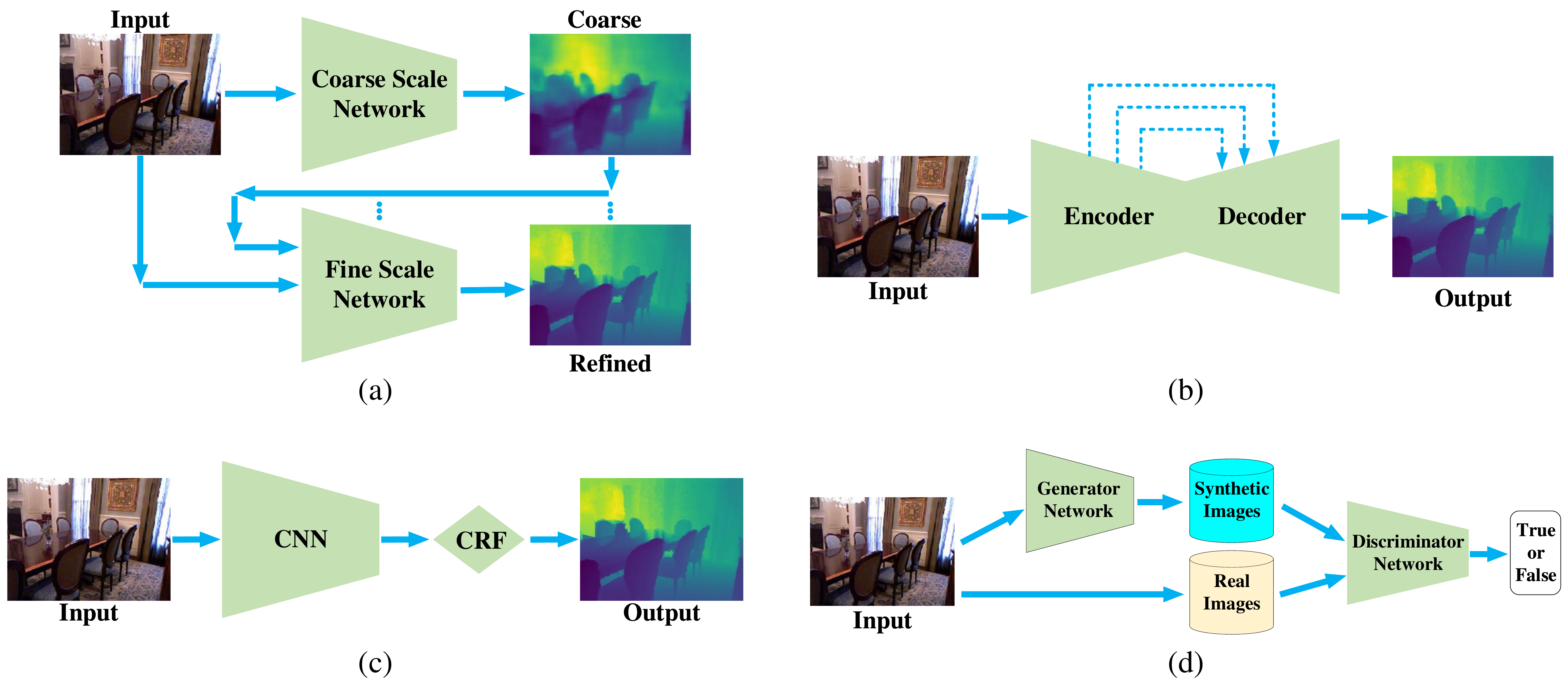} \\
	\caption{Taxonomy of different network architectures. (a) Multi-scale Network \cite{eigen2014depth, eigen2015predicting}, (b) Encoder-Decoder Network (dotted lines represent skip connections) \cite{laina2016deeper, ma2018sparse, cheng2018depth, hu2019revisiting, chen2019structure}, (c) CNN Combines with CRF \cite{li2015depth, cao2017estimating, heo2018monocular}, (d) GANs \cite{aleotti2018generative, mehta2018structured}. (Best viewed in color).}
	\label{fig:networks}
\end{figure*}
\begin{figure*}[tb]
	\centering
	\includegraphics[width=.9\linewidth]{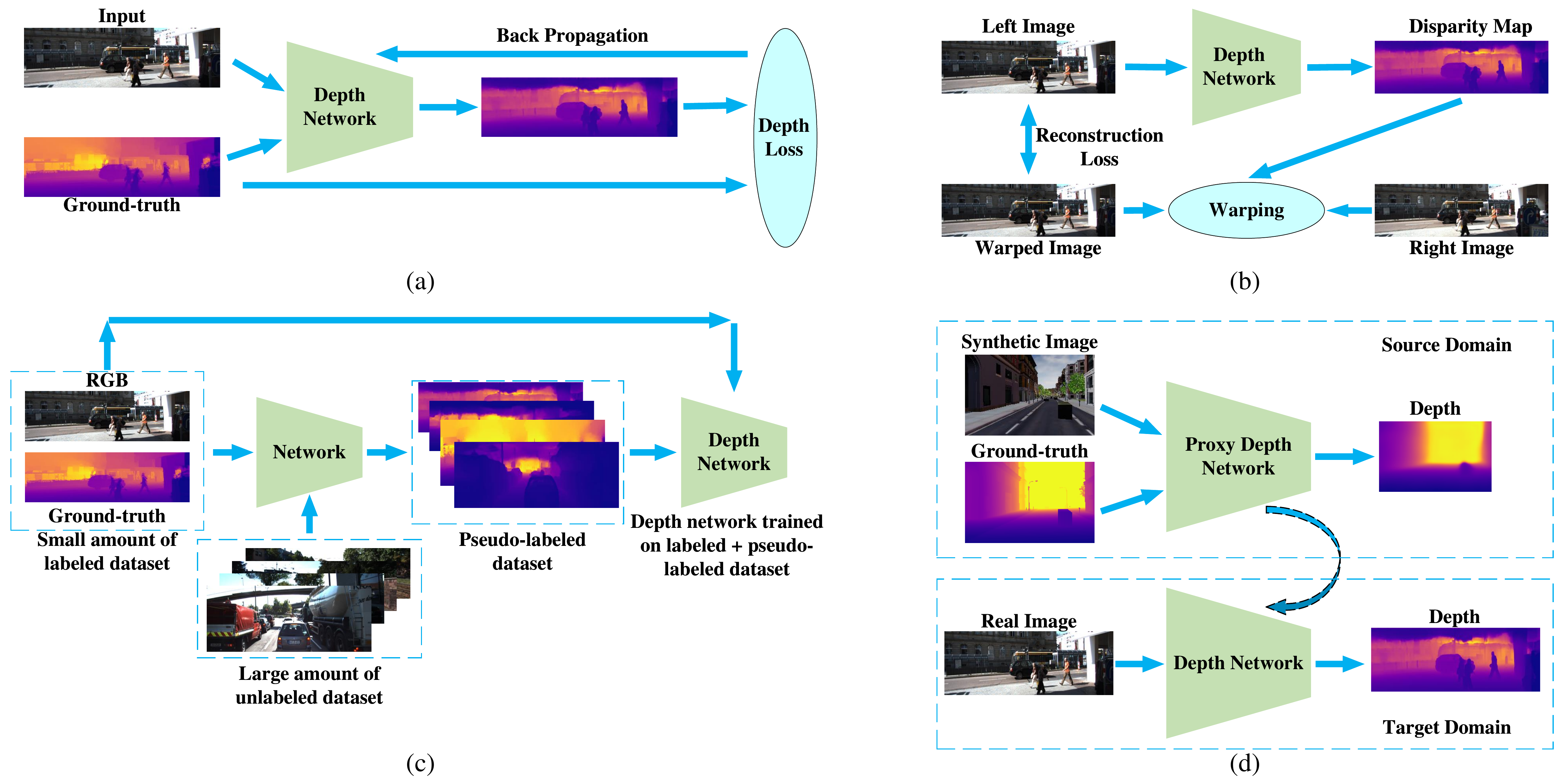} \\
	\caption{The general architecture of deep learning-based monocular depth estimation. (a) Supervised learning network, which takes an RGB image and ground-truth depth as input and outputs the estimated depth map, (b) Unsupervised learning network takes as input the stereo image, (c) Semi-supervised learning network, which uses a small amount of image-depth pairs and a large amount of unlabeled images, and (d) Domain adaptation method, where the network in the target domain is trained on synthetic data, the arrow with dotted line represents adaptation. (Best viewed in color).}
	\label{fig:schemes}
\end{figure*}
\subsection{Depth Estimation with Supervised Learning} \label{subsection: supervised}
The pipeline of supervised learning-based MDE methods can be described as follows (see Fig. \ref{fig:schemes} (a)). The MDE network incorporates a single image $I$ and the corresponding ground-truth depth map $D^{*}$ to learn the scene structure information for estimating a dense depth map ($D$). Then the parameters of the network is updated by minimizing a loss function $L(D^{*}, D)$, which measures the difference between $D$ and $D^{*}$. The network converges when $D$ is as close as possible to $D^{*}$. \par
\subsubsection{General Supervised Methods}
The general supervised methods treat MDE as a regression problem. To our knowledge, Eigen et al. \cite{eigen2014depth} introduced the first deep learning-based MDE algorithm. In order to exploit global and local information, two CNNs are employed in this work (Fig \ref{fig:networks}(a)). In addition to the common scale-dependent errors, a scale-invariant error is used as the loss function to optimize the training. The real scale of depth is recovered without any post-processing. This work remarkably improved the performance of MDE on the NYU \cite{silberman2012indoor} and KITTI \cite{geiger2013vision} datasets. Considering the continuous nature of depth values, Liu et al. \cite{liu2015learning} cast depth estimation as a deep continuous conditional random fields (CRF) learning problem. They design a network which includes three modules: unary part, pairwise part and continuous CRF loss layer. The input images are first segmented into superpixels. Only image patches centered around each super-pixel are passed to the designed network to predict depth values. \cite{eigen2014depth, liu2015learning} depend on fully connected (FC) layers to predict depth values. While the FC layer yields a full receptive field, it has a huge number of trainable parameters resulting in \cite{liu2015learning} needing over a second to estimate a single depth map from a test image. \par 
Laina et al. \cite{laina2016deeper} introduce a fully convolutional residual network (FCRN) for depth estimation. FCRN consists of two parts, encoder and decoder (Fig \ref{fig:networks}(b)). The encoder is modified from ResNet-50 \cite{he2016deep} by removing the FC layers and the last pooling layer. The decoder guides the network into learning its upscaling via a series of upsample and convolutional layers. The removal of FC layers significantly reduces the number of learnable parameters. Their experiments demonstrate that with the increase of network depth, the accuracy apparently increases, because a deeper network leads to a larger receptive field and captures more context information. Inspired by this finding, CNNs with more than 100 layers e.g., ResNet-101/152 \cite{he2016deep}, DenseNet-169 \cite{huang2017densely} or SENet-154 \cite{hu2018squeeze} have been applied to MDE. Mancini et al. \cite{mancini2016fast} utilize optical flow as additional information to estimate depth from monocular images. They concatenate the current RGB image with an optical flow map generated from the current frame and the previous one. The stacked images are fed to an encoder-decoder architecture to learn depth from fused information. \par
Based on DenseNet-169 \cite{huang2017densely}, Alhashim and Wonka \cite{alhashim2018high} design a densely connected encoder-decoder architecture. Unlike \cite{laina2016deeper}, they use a simple decoder method which consists of a bilinear upsampling and two convolution layers. With the deeper network architecture, elaborated augmentation and training strategies, the designed network generates more accurate results on the NYU \cite{silberman2012indoor} and KITTI \cite{geiger2013vision} datasets. To effectively guide the mapping from the densely extracted features to the desired depth estimation, Lee et al. \cite{lee2019big} design a local planar guidance (LPG) layer and apply it to each decoding stage. The output from the LPG layers has the same size as the desired depth map. Then, the outputs are combined to get the final estimation. Yin et al.\cite{yin2019enforcing} construct a geometric constraint in the 3D space for depth estimation by designing a loss term. The designed loss function combines geometric with pixel-wise depth supervisions, which enables the depth estimation network generates accurate depth map and high quality 3D point cloud. \par
Hu et al. \cite{hu2019revisiting} combine an encoder-decoder module with a multi-scale feature fusion (MSFF) module and a refinement (RF) module. The MSFF module upscales feature maps from different encoder layers to the same size and then concatenate it channel by channel. Features from the MSFF module are combined with features from the decoder and then fed to the RF module to generate the final prediction. The main contribution of \cite{hu2019revisiting} is a hybrid loss function, which measures errors in depth, gradients and surface normals. Inspired by \cite{hu2019revisiting}, Chen et al. \cite{chen2019structure} design a Structure-Aware Residual Pyramid Network (SARPN) to exploit scene structures in multiple scales for MDE. SARPN includes three parts, an encoder which extracts multi-scale features, an adaptive dense feature fusion module for dense feature fusion and a residual pyramid decoder. The residual pyramid decoder estimates depth maps at multiple scales to restore the scene structure in a coarse-to-fine manner. \par
Fu et. al. \cite{fu2018deep} discretize the continuous depth into a series of intervals and transfer depth estimation to an ordinal regression problem. As the uncertainty of the estimated depth increases along with the ground-truth depth values, the common uniform discretization (UD) strategy may result in an over-strengthened loss for the large depth values. In order to solve this problem, a spacing increasing discretization (SID) strategy is designed to discretize the depth values. With the obtained discrete depth values, an ordinal regression loss which involves the ordered information between discrete labels is applied to train the network. \par
Bhat et al. \cite{bhat2021adabins} divide the estimated depth range into bins where the bin widths change per image. This enables the network to learn to adaptively focus on the regions of different depths. The main contribution of \cite{bhat2021adabins} is a Mini-ViT module consisting of four transformer layers \cite{vaswani2017attention}. Being designed as a variant of Vision Transformer \cite{dosovitskiy2020image}, Mini-ViT takes as input the multi-channel feature map of the input image to compute global information at a high resolution and outputs bin-widths and range-attention-maps showing the likelihood of each bin. Unlike Fu et al. \cite{fu2018deep} estimate depth as the bin center of the most likely bin. The final depth of \cite{bhat2021adabins} is the linear combination of bin centers weighted by the probabilities. Hence, this approach generates smoother depth maps. \par
In recent years, omnidirectional cameras have become more and more popular. Depth estimation from single $360^{\circ}$ images \cite{zioulis2018omnidepth, wang2020bifuse} also being explored by researchers. Compared with the regular cameras, omnidirectional cameras have a larger field-of-view which enables them to record the entire surroundings. According to Google Scholar, the first work to estimate a depth map from an omnidirectional image is OmniDepth \cite{zioulis2018omnidepth}. The main contribution of \cite{zioulis2018omnidepth} is a dataset consisting of $360^{\circ}$ RGB-depth pairs. Since acquiring $360^{\circ}$ datasets with ground-truth is difficult, the authors resort to re-use recently released 3D datasets to produce diverse $360^{\circ}$ view images. Wang et al. \cite{wang2020bifuse} propose a two-branch framework which combines equirectangular and cubemap projection to infer depth from monocular $360^{\circ}$ images. The two branches take equirectangular image and cubemap as input, respectively. The produced features are combined by a bi-projection fusion block to exploit the shared feature representations. \par
%
In addition to Convolutional Neural Networks (CNNs), Recurrent Neural Networks (RNNs) are also applied to the application of MDE. RNNs are a class of neural networks that model the temporal behavior of sequential data through hidden states with cyclic connections. Unlike CNNs that back-propagate gradient through the network, RNNs additionally back-propagate the gradient through time. Therefore, RNNs can learn dependencies across time. As an extension of the regular RNN, the long short-term memory (LSTM) is able to learn long-term dependencies within the input sequence. \par
Kumar et al. \cite{cs2018depthnet} design a convolutional LSTM (ConvLSTM) based encoder-decoder architecture to learn depth from the spatio-temporal dependencies between video frames. The encoder consists of a set of ConvLSTM layers and the decoder includes a sequence of deconvolutional and convolutional layers. Each ConvLSTM layer has $N$ states that correspond to the number of timestamps, thus, the network learns depth maps from $N$ consecutive video frames. Zhang et al. \cite{zhang2019exploiting} exploit spatial and temporal information for depth estimation by combining ConvLSTM and Generative Adversarial Network (GAN). In addition, a temporal consistency loss is designed to further maintain the temporal consistency among video frames. The designed temporal loss is combined with the spatial loss to update the model in an end-to-end manner. \par
A RNN-based multi-view method for learning depth and camera pose from monocular video sequences is introduced by \cite{wang2019recurrent}. The ConvLSTM units are interleaved with convolutional layers to exploit multiple previous frames in each estimated depth map. With the model of multi-views, the image reprojection constraint between multi view images can be incorporated into the loss function. Additionally, a forward-backward flow-consistency constraint is applied to solve the ambiguity of image reprojection by providing additional supervision.
\subsubsection{Monocular Depth Estimation by Classification}
For different pixels in a single image, the possible depth values may have different distributions. Therefore, depth estimation can be formulated as a pixel-wise classification task by discretizing the continuous depth values into segments \cite{cao2017estimating, li2018deep, li2018monocular, chen2019attention, zou2019mean}. Cao et al. \cite{cao2017estimating} design a fully convolutional deep residual network to estimate the depth range. Li et al. \cite{li2018monocular} propose a hierarchical fusion dilated CNN to learn the mapping between the input RGB image and corresponding depth map. A soft-weighted-sum inference is proposed to transfer the discretized depth scores to continuous depth values. \par 
Following \cite{cao2017estimating, li2018deep, li2018monocular}, Zou et al. \cite{zou2019mean} cast depth estimation as a classification problem but take probability distribution into account in the training step. The main contribution of \cite{zou2019mean} is a novel mean-variance loss which consists of a mean loss and a variance loss. The mean loss is used to penalize the error between the mean of estimated depth distribution and the ground-truth. Meanwhile the variance loss is complementary to the mean loss and makes the distribution sharper. The mean-variance loss is combined with the softmax loss to supervise the training of the depth estimation network. \par
In addition, depth estimation can also be solved by combining depth regression and depth interval classification together \cite{liebel2019multidepth, song2019contextualized}. Song et al. \cite{song2019contextualized} exploit the shared features from the semantic labels, contextual relations and depth information in a unified network. They use a FCN-based network to encode the input RGB images into high-level semantic feature maps. The generated feature maps are passed to a two step decoder. In the first step, the feature maps are sampled and fed into a ``semantic decoder" to up-sample semantic class labels, and a ``depth decoder" to estimate the depth map. Subsequently, the semantic map and depth map are refined at the class and pixel level by a CRF layer. At last, the classification and regression tasks are integrated to model the depth estimation process by a joint-loss layer and produce the final depth map. \par
\subsubsection{Multi-task Learning Based Methods}
Depth estimation and other applications such as semantic segmentation and surface normal estimation are correlated and mutually beneficial. For example, semantic maps and depth maps reveal the layout and object boundaries/shapes \cite{sanchez2018hybridnet}. In order to take advantage of the complementary properties of these tasks, multi-task learning in a unified framework has been explored \cite{eigen2015predicting, wang2015towards, jafari2017analyzing, gurram2018monocular, jiao2018look, qi2018geonet, nekrasov2019real, hsieh2019deep, abdulwahab2020adversarial}. \par
Eigen and Fergus \cite{eigen2015predicting} design a unified three-scale network for three different tasks, depth estimation, surface normal estimation and semantic segmentation. The first scale block estimates a coarse global feature from the entire image. Then the feature map is passed to the second and third scale blocks. The second scale block produces mid-level resolution estimations, and the third scale outputs higher resolution estimations at half size of the input image. The designed network can be trained for three different tasks by changing the output layer and the loss function. \par 
Inspired by the global network of \cite{eigen2014depth}, Wang et al. \cite{wang2015towards} propose a CNN that jointly estimates pixel-wise depth values and semantic labels. To obtain fine-level details, the authors decompose input images into local segments and use the global layouts to guide the estimation of region-level depth and semantic labels. With the global and local estimations, the inference problem is formulated to a two-layer Hierarchical Conditional Random Field to produce the refined depth and semantic map. Jafari et al. \cite{jafari2017analyzing} design a modular CNN that jointly solve MDE and semantic segmentation. The designed network consists of an \emph{estimation} module and a \emph{refine} module. The estimation module utilizes \cite{eigen2015predicting} and \cite{long2015fully} as sub-networks for different tasks respectively. During training, the two sub-networks positively enforce each other and mutually improve each other. The refine module incorporates the output of the estimation module and produces refined estimations. \par
The above-mentioned methods \cite{eigen2015predicting, wang2015towards, jafari2017analyzing} require training images have per-pixel labels with depth and semantic class ground-truth. It is difficult to collect such datasets, especially for outdoor scenarios. Gurram et al. \cite{gurram2018monocular} solve this problem by leveraging depth and semantic information from two heterogeneous datasets to train a depth estimation CNN. The training process is divided into two steps. In the first step, a multi-task learning scheme is applied for pixel-level depth and semantic classification. In the second step, the regression layers take as input the classified depth maps in order to generate the final depth maps. \par
Qi et al. \cite{qi2018geonet} design a Geometric Neural Network (GeoNet) that jointly learns depth and surface normals from monocular images. Being designed as a two-stream CNN, GeoNet consists of a depth-to-normal (DTN) and a normal-to-depth (NTD) network. The DTN network infers surface normal from depth map via the least square solution and a residual module, while the NTD network refines the depth estimation from the estimated surface normal and initial depth map. Hesieh et al. \cite{hsieh2019deep} propose a multi-task learning network by adding a depth estimation branch to YOLOv3 \cite{redmon2018yolov3}. In the process of training, an $L_{1}$ distance depth estimation loss ($\sum_{i}^{N}|depth_{i} - depth_{i}^{*}|$, where $N$ is the amount of objects in a batch, $depth_{i}$ and $depth_{i}^{*}$ denote the estimated and ground-truth object depth) is directly added to the object detection loss function. \par
Abdulwahab et al. \cite{abdulwahab2020adversarial} introduce a framework for predicting the depth and 3D pose of the main objects shown in the input image. The proposed framework stacks a GAN block and a regression CNN block in series connection. The GAN block is trained with a loss function for feature matching which enables the network to generate a dense depth map from an input image, while the regression block incorporates the generated depth map to predict the 3D pose. The supervised multi-task learning methods estimate depth maps with other tasks such as semantic estimation and surface normal estimation can improve the accuracy of the depth map. However, it is difficult to collect datasets with depth labels and other labels. \par
\subsubsection{Real-Time Supervised Monocular Depth Estimation} The aforementioned algorithms are based on complex DNNs which are challenging for real-time requirements. In order to enable MDE network to run at real-time speed on embedded platforms, Spek et al. \cite{spek2018cream} build a lightweight depth estimation network on top of the ``non-bottleneck-1D" block \cite{romera2017erfnet}. The designed network runs about 30fps on the Nvidia-TX2 GPU, but its accuracy is inferior. \par
Later, Wofk et al. \cite{wofk2019fastdepth} develop a lightweight encoder-decoder network for monocular depth estimation. In addition, a network pruning algorithm is applied to further reduce the amount of parameters. Experimental results on the NYU dataset show that the obtained depth estimation network runs at 178 fps on an Nvidia-TX2 GPU, while the RMSE and $\delta_{1}$ values are 0.604 and 0.771 respectively. Inspired by the densely-connected encoder-decoder architecture \cite{alhashim2018high}, Wang et al. \cite{wang2020depthnet} design a highly compact network named DepthNet Nano. DepthNet Nano applies densely connected projection batchnorm expansion projection (PBEP) modules to reduce network architecture and computation complexity while maintaining the representative ability. \par
Supervised learning methods require vast amounts of depth images as ground-truth for training, allowing these methods to achieve high accuracy for MDE. However, collecting this ground-truth data from the real world requires depth sensing devices such as LIDAR or RGB-D cameras, which increases the expense. In addition, these sensors require accurate extrinsic and intrinsic calibration, any error in calibration results in an inaccurate ground-truth. Therefore, unsupervised learning-based methods \cite{garg2016unsupervised, godard2017unsupervised, zhou2017unsupervised} which do not require ground-truth are attracting attention. \par
\subsection{Depth Estimation with Unsupervised Learning}
{Unsupervised learning methods take as input stereo images or video sequences with small changes in camera positions between frames as two continuous frames can be treated as stereo images. These methods formulate depth estimation as an image reconstruction problem, where depth maps are an intermediate product that integrates into the image reconstruction loss. The pipeline of unsupervised learning-based methods (see Fig. \ref{fig:schemes}(b)) can be described as follows: the network incorporates two images (name it $I_{L}$ and $I_{R}$) of the same scene but with slightly different perspectives. Subsequently, the depth map is estimated for $I_{L}$, and the obtained depth map is represented by $D_{L}$. With $D_{L}$ and the camera motion between images, $I_{R}$ can be warped to an image which is similar to $I_{L}$ through Equation \ref{eq:wrap}.
\begin{equation} \label{eq:wrap}
	I_{R}(D_{L}) \to \tilde{I}_{L}
\end{equation}
where $\tilde{I}_{L}$ is the warped image. The network can be trained with the reconstruction loss formulated in Equation \ref{eq:reconstruction_loss}}. 
\begin{equation}\label{eq:reconstruction_loss}
	loss = L({\tilde{I}_{L}}, {I_{L}})
\end{equation} 
\subsubsection{General Unsupervised Methods}
According to our literature review, Garg et al. \cite{garg2016unsupervised} developed the first unsupervised learning method for MDE. In this work, image pairs with known camera motion are fed to the network to learn the non-linear transformation between the source image and depth map. The color constancy error between the input image and the inverse-warped target image is used as the loss to optimize the update of network weights. In addition to camera motion, the correspondence between the left and right images are also cues for unsupervised MDE. Godard et al. \cite{godard2017unsupervised} train an encoder-decoder network in an unsupervised manner by designing a left-right consistency loss. With the calibrated stereo image pairs and epipolar geometry constraints, the designed method does not need depth data as a supervisory signal. Both \cite{garg2016unsupervised} and \cite{godard2017unsupervised} require calibrated stereo pairs for training. Thus, datasets without stereo images \cite{saxena2008make3d, silberman2012indoor} cannot be applied to train these methods. \par 
Mahjourian et al. \cite{mahjourian2018unsupervised} alleviate the dependence on stereo images by exploiting the consistency between depth and ego-motion from continuous frames as supervisory signal for training. Tosi et al. \cite{tosi2019learning} propose an unsupervised framework that infers depth from a single input image by synthesizing features from a different point of view. The designed network includes three parts: multi-scale feature extractor, initial disparity estimator and disparity refinement module. Given an input image, high-level features at different scales are extracted by the multi-scale feature extractor. The extracted features are passed to the initial disparity estimator to predict multi-scale disparity maps that aligned with the input and synthesized right view image. The refinement module refines the initial disparity by performing stereo matching between the real and synthesized feature representations. \par
Ma et al. \cite{ma2019self} extend \cite{ma2018sparse} to an unsupervised approach. The input sparse depth maps and the RGB images are pre-processed by initial convolutions separately. The output features are concatenated into a single tensor, which are passed to the encoder-decoder framework. The network is trained in an unsupervised scheme. Besides, the authors use the Perspective-n-Pose method to estimate pose, which assumes that the input sparse depth is noiseless and susceptible to failure in low image texture situations. Based on \cite{ma2019self}, Zhang et al. \cite{zhang2019dfinenet} design a framework that jointly learns depth and pose from monocular images. The temporal constraint is applied to measure the reprojection error and provides a training signal to depth and pose CNNs simultaneously. The reprojection error signal works on the noisy sparse depth input, while the ground-truth depth provides scale information and supervises the training of depth estimation. \par 
Fei et al. \cite{fei2019geo} design an unsupervised network that uses the global orientation and the semantics of the scene as the supervisory signal. Unlike previous work that computes the surface normals from the depth values first and then impose regularity, they directly regularize the depth values via the scale-invariant constraint. Guizilini et al. \cite{guizilini2020semantically} utilize semantic information to guide the geometric representation learning of MDE. The designed architecture is built within an unsupervised scheme \cite{guizilini20203d}. It consists of two networks, one responsible for depth estimation whilst the other performs semantic segmentation. During training, only the depth estimation network is optimized, while the weights of the semantic segmentation network are fixed to guide the depth estimation network to learn features via pixel-adaptive convolutions. Recently, Johnston and Carneiro \cite{johnston2020self} introduce a discrete disparity volume to regularize the training of an unsupervised network. The designed method enables the network to predict sharper depth map and pixel-wise depth uncertainties. \par
\subsubsection{Multi-task Learning Based Methods}
Zhou et al. \cite{zhou2017unsupervised} present a method that jointly learning depth maps and camera motion from monocular videos. The proposed framework stacks a depth network \cite{mayer2016large} and a pose network. The produced depth maps and relative camera pose are applied to inverse warp the source views to reconstruct the target view. By using view synthesis as the supervisory signal, the entire framework can be trained in an unsupervised manner. Due to the dependence on two frame visual odometry estimation method, this network suffers from the per frame scale ambiguity problem. \par
Inspired by \cite{zhou2017unsupervised}, Prasad and Bhowmich \cite{prasad2019sfmlearner++} use epipolar constrains to optimize the joint learning of depth and ego-motion. The main idea behind the training is similar to \cite{mayer2016large}. Instead of using epipolar constrains as labels for training, the authors apply it to weight the pixels to guide the training. Klodt and Vedaldi \cite{klodt2018supervising} modify \cite{zhou2017unsupervised} in the following aspects. Firstly, a structural similar loss is imported to strengthen the brightness constancy loss. Besides, an explicit model of confidence is incorporated to the network by predicting each pixel a distribution over possible brightnesses. Finally, a SfM algorithm \cite{mur2017orb} is applied to the network to provide supervisory signal for the training of depth estimation network. \par
Vijayanarasimhan et al. \cite{vijayanarasimhan2017sfm} design ``SfM-Net," a geometry-aware network capable of estimating depth, camera motion and dynamic object segmentation. The designed network includes two sub-networks, the structure network learns to estimate depth while the motion network predicts camera and object motion. The outputs from both sub-networks are then transformed into optical flow by projecting the point cloud from depth estimation to the image space. Thus, the network can be trained in an unsupervised manner through minimizing the photometric error. Dai et al. \cite{dai2019self} propose a self-supervised learning framework for jointly estimating individual object motion and depth from monocular video. Instead of modeling the motion by 2D optical flow or 3D scene flow, the object motion is modeled and predicted in the form of full 6 degrees of freedom (6 DoF). \par 
Joint learning of depth estimation and pose estimation is usually done under the assumption that a consistent scale of CNN-based MDE and relative pose estimation can be learned across all input samples. This hypothesis degrades the performance in environments where the changes of relative pose across sequences are significantly remarkable. In order to tackle the problem of scale inconsistency, Bian et al. \cite{bian2019unsupervised} design a geometry consistency loss as shown in Equation \ref{eq:l_gc}. With the proposed loss function, the depth and ego-motion networks are trained in monocular videos to predict scale-consistent results. Given any two continuous images $(I_a, I_b)$ from an unlabeled video, they first use a depth network to compute the corresponding depth maps $(D_a, D_b)$, and then compute the relative 6 DoF pose $P_{ab}$ between them using a pose network. With the obtained depth and relative camera pose, the warped $D^{a}_{b}$ is computed by transforming $D_a$ to 3D space and projecting it to $I_b$ using $P_{ab}$. The inconsistency between $D^{a}_{b}$ and $D^{'}_{b}$ are used as geometric consistency loss $L_{GC}$ to supervise the training of the network. {$L_{GC}$ is defined as in Equation \ref{eq:l_gc}.}
\begin{equation} \label{eq:l_gc}
L_{GC} = \frac{1}{|V|}{\sum_{p \in V}}{D_{diff}(p)}
\end{equation}
where $V$ represents valid points that are successfully projected from $I_{a}$ to the image plane of $I_{b}$, $\lvert V \rvert$ means the number of points in $V$, and $D_{diff}$ stands for the depth inconsistency map. For each point $p$ in $V$, $D_{diff}$ is defined in Equation \ref{eq:d_diff}.  
\begin{equation} \label{eq:d_diff}
	 {D_{diff}(p)} = \frac{|{D_{b}^{a}(p) - D_{b}^{'}(p)|}}{D_{b}^{a}(p) + D_{b}^{'}(p)}
\end{equation}
In order to mitigate the influence of moving objects and occlusions on network training, Bian et al. design a self-discovered mask $M$ ($M = 1 - D_{diff}$) which assigns low/high weights for inconsistent/consistent pixels. \par
Zhao et al. \cite{zhao2020towards} disentangle scale from the joint learning of depth and relative pose. Unlike \cite{zhou2017unsupervised} and \cite{prasad2019sfmlearner++} utilize PoseNet \cite{kendall2015posenet} to generate relative pose, the work in \cite{zhao2020towards} directly predicts relative pose by solving the fundamental matrix from dense optical flow correspondence and apply a differentiable two-view triangulation module to recover an up-to-scale 3D structure. The depth error is measured after a scale adaptation from the estimated depth to the triangulated structure and the reprojection error between depth and flow is computed to further enforce the end-to-end joint training. \par
Zou et al. \cite{zou2018df} present an unsupervised framework to jointly learn depth and optical flow from monocular video sequences. In addition to the regular photometric and spatial smoothness loss, a cross-task consistency loss is designed to provide additional supervisory signals for both tasks. Yin and Shi \cite{yin2018geonet} jointly learn depth, optical flow and camera pose in a unified network. They use a rigid structure reasoning module to infer scene architecture, and a non-rigid motion refinement module to cope with the effect of dynamic objects. These two modules work in different stages, and the view synthetics are used as a basic supervision for the unsupervised learning paradigm. Furthermore, an adaptive geometric consistency loss is designed to tackle the occlusions and texture ambiguities that is not included in pure view synthesis objectives. \par
Ranjan et al. \cite{ranjan2019competitive} learn depth along with camera motion estimation, optic flow estimation and motion segmentation. In order to achieve the goal of joint learning, they design a Competitive Collaboration (CC) learning method. It consists of two modules, the static scene reconstructor infers the static scene pixels using depth and camera motion, and the moving region reconstructor reasons about pixels in the independently moving regions. The two modules compete for a resource whilst being regulated by a moderator, the motion segmentation network. The CC method coordinates the training of multiple tasks and achieves performance gains in both tasks. \par
%
\subsubsection{Adversarial Learning Based Methods}
In addition to learning depth from view-synthesis or minimizing photometric reconstruction error, unsupervised MDE \cite{aleotti2018generative, mehta2018structured, pilzer2018unsupervised, wang2020adversarial, almalioglu2019ganvo} has also been solved by Generative Adversarial Networks (GANs). GANs consist of a generator network and a discriminator network (Fig. \ref{fig:networks}(d)). The two networks are trained by the backpropagation algorithm, thus they can work together to construct unsupervised learning models. Since there is no ground-truth depth in unsupervised learning, the discriminator distinguishes between the synthesized and the real images. \par
Aleotti et al. \cite{aleotti2018generative} present the first generative adversarial network for unsupervised MDE. The generator network is trained to infer a depth map from the input image to generate a warped synthesized image. The discriminator network is trained to distinguish the warped image and the input real image. Since the quality of the estimated depth maps has an effect on the warped synthesized images, the generator is forced to generate more accurate depth maps. Mehta et al. \cite{mehta2018structured} introduce a structural adversarial training method which predicts dense depth maps using stereo-view synthesis. Given a monocular image, the generator network outputs a dense disparity maps. With the produced disparity map, multi-view stereo pairs corresponding to the input image view are generated. The discriminator network distinguishes these reconstructed views from the real views in the training data. \par
Wang et al. \cite{wang2020adversarial} integrate adversarial learning with spatial-temporal geometric constraints for the joint learning of depth and ego-motion. The generator combines depth-pose net with direct visual odometry {DVO} to produce a synthesized image. The combination of posenet and DVO generates a fine-grained pose estimation and provides an effective back-propagation gradient to the depth network. Meanwhile, the discriminator takes the synthesized and original images to distinguish the reconstructed and real images. Almalioglu et al. \cite{almalioglu2019ganvo} design an adversarial and recurrent unsupervised learning framework. The designed network consists of a depth generator and a pose regressor. With the produced depth map, 6 DoF camera pose and color values from the source images, the view reconstruction module synthesizes a target image. The discriminator network distinguishes the synthesized target image from the real target image. \par
\subsubsection{Real-Time Unsupervised Monocular Depth Estimation} Although these works achieve promising performance, however, they all have fairly deep and complex architectures. Therefore, real-time speed can only be achieved on high-performance GPUs, which inhibits their application in autonomous driving or robotics. In order to tackle the problem of running speed, Poggi et al. \cite{poggi2018towards} stack a simple encoder and multiple small decoders working in a pyramidal structure. The designed network only has 1.9M parameters and requires 0.12s to produce a depth map on a i7-6700K CPU, which is close to a real-time speed. Liu et al. \cite{liu2020mininet} introduce a lightweight model (named MiniNet) trained on monocular video sequences for unsupervised depth estimation. The core part of MiniNet is DepthNet, which iteratively utilizes the recurrent module-based encoder to extract multi-scale feature maps. The obtained feature maps are passed to the decoder to generate multi-scale disparity maps. MiniNet achieves real-time speed about 54fps with $640\times192$ sized images on a single Nvidia 1080Ti GPU. \par
Unsupervised learning methods formulate MDE as an image reconstruction problem and use geometric constraints as supervisory signal. This category of methods take stereo images or monocular image sequences as input to learn geometry constraints between the left and right images or continuous frames. Unsupervised learning methods do not require ground-truth in the training process, which avoids the expense of collecting depth maps. However, due to the absence of ground-truth the accuracy rate is inferior to supervised learning methods (see Table \ref{table: comparison_of_mde_on_kitti}). 
\begin{table*}[!tb]
	\caption{A summary of deep learning-based monocular depth estimation algorithms with open-source implementations. ``S": supervised, ``U": unsupervised, ``Semi": semi-supervised and ``D" : domain adaptation.}
	\renewcommand{\arraystretch}{1.3}
	\label{table: implementations_of_mde}
	\centering
	\begin{tabular}{c | c | c | c | c} \hline\hline
		Year & Algorithm & Type & Implementation & Source code \\ \hline\hline
		2014 & Eigen et al. \cite{eigen2014depth} & S & Python & https://cs.nyu.edu/~deigen/depth/ \\ \hline
		2015 & Eigen et al. \cite{eigen2015predicting} & S & Python & https://cs.nyu.edu/~deigen/dnl/ \\ \hline
		2016 & Laina et al. \cite{laina2016deeper} & S & \makecell{TensorFlow, MatConvNet} & https://github.com/iro-cp/FCRN-DepthPrediction \\ \hline
		2017 & Xu et al. \cite{xu2017multi} & S & Caffe & https://github.com/danxuhk/ContinuousCRF-CNN.git \\ \hline
		2018 & Alhashim and Wonka \cite{alhashim2018high} & S & \makecell{PyTorch, TensorFlow} & https://github.com/ialhashim/DenseDepth \\ \hline		
		2018 & Fu et al. \cite{fu2018deep} & S & Caffe & https://github.com/hufu6371/DORN \\ \hline
		2018 & Guo et al. \cite{guo2018learning} & S & PyTorch & https://github.com/xy-guo/Learning-Monocular-Depth-by-Stereo \\ \hline
		2018 & Li and Snavely \cite{li2018megadepth} & S & PyTorch & https://github.com/zhengqili/MegaDepth \\ \hline
		2018 & Ma and Karaman \cite{ma2018sparse} & S & PyTorch & https://github.com/fangchangma/sparse-to-dense.pytorch \\ \hline		
		2018 & Zioulis et al. \cite{zioulis2018omnidepth} & S & PyTorch & https://github.com/VCL3D/SphericalViewSynthesis \\ \hline
		2019 & Bian et al. \cite{bian2019unsupervised} & S & PyTorch & https://github.com/JiawangBian/SC-SfMLearner-Release \\ \hline
		2019 & Chen et al. \cite{chen2019structure} & S & PyTorch & https://github.com/Xt-Chen/SARPN \\ \hline
		2019 & Hu et al. \cite{hu2019revisiting} & S & PyTorch & https://github.com/JunjH/Revisiting-Single-Depth-Estimation \\ \hline				
		2019 & Lee et al. \cite{lee2019big} & S & \makecell{PyTorch, TensorFlow} & https://github.com/cogaplex-bts/bts \\ \hline
		2019 & Liebel and Ko¨rner \cite{liebel2019multidepth} & S & Pytorch & https://github.com/lukasliebel/MultiDepth \\ \hline
		2019 & Nekrasov et al. \cite{nekrasov2019real} & S & PyTorch & https://github.com/DrSleep/multi-task-refinenet \\ \hline		
		2019 & Qiu et al. \cite{qiu2019deeplidar} & S & PyTorch & https://github.com/JiaxiongQ/DeepLiDAR \\ \hline   
		2019 & Wofk et al. \cite{wofk2019fastdepth} & S & PyTorch & https://github.com/dwofk/fast-depth \\ \hline
		2019 & Yin et al. \cite{yin2019enforcing} & S & Pytorch & https://tinyurl.com/virtualnormal \\ \hline
		2020 & Fang et al. \cite{fang2020towards} & S & PyTorch & https://github.com/zenithfang/supervised-dispnet \\ \hline
		2020 & Sartipi et al. \cite{sartipi2020deep} & S & PyTorch & https://github.com/MARSLab-UMN/vi-depth-completion \\ \hline
		2020 & Xian et al. \cite{xian2020structure} & S & PyTorch & https://github.com/KexianHust/Structure-Guided-Ranking-Loss \\ \hline 
		2021 & Bhat et al. \cite{bhat2021adabins} & S & Pytorch & https://github.com/shariqfarooq123/AdaBins \\ \hline \hline
		2016 & Garg et al. \cite{garg2016unsupervised} & U & Caffe & https://github.com/Ravi-Garg/Unsupervised-Depth-Estimation \\ \hline
		2017 & Godard et al. \cite{godard2017unsupervised} & U & TensorFlow & https://github.com/mrharicot/monodepth \\ \hline
		2017 & Zhou et al. \cite{zhou2017unsupervised} & U & TensorFlow & https://github.com/tinghuiz/SfMLearner \\ \hline 
		2018 & Pilzer et al. \cite{pilzer2018unsupervised} & U & TensorFlow & https://github.com/andrea-pilzer/unsup-stereo-depthGAN \\ \hline
		2018 & Poggi et al. \cite{poggi2018towards} & U & TensorFlow & https://github.com/mattpoggi/pydnet \\ \hline
		2018 & Qi et al. \cite{qi2018geonet} & U & TensorFlow & https://github.com/xjqi/GeoNet \\ \hline		
		2018 & Zhan et al. \cite{zhan2018unsupervised} & U & Caffe2 & https://github.com/Huangying-Zhan/Depth-VO-Feat \\ \hline
		2019 & Casser et al. \cite{casser2019depth} & U & TensorFlow & \makecell{https://github.com/tensorflow/models/tree/archive/research/struct2depth} \\ \hline 		
		2019 & Elkerdawy et al. \cite{elkerdawy2019lightweight} & U & TensorFlow & https://github.com/selkerdawy/joint-pruning-monodepth \\ \hline		
		2019 & Fei et al. \cite{fei2019geo} & U & TensorFlow & https://github.com/feixh/GeoSup \\ \hline
		2019 & Godard et al. \cite{godard2019digging} & U & PyTorch & https://github.com/nianticlabs/monodepth2 \\ \hline						 		
		2019 & Ranjan \cite{ranjan2019competitive} & U & PyTorch & https://github.com/anuragranj/cc \\ \hline 
		2019 & Tosi et al. \cite{tosi2019learning} & U & TensorFlow & https://github.com/fabiotosi92/monoResMatch-Tensorflow \\ \hline		
		2019 & Watson et al. \cite{watson2019self} & U & PyTorch & https://github.com/nianticlabs/depth-hints \\ \hline   
		2019 & Zioulis et al. \cite{zioulis2019spherical} & U & PyTorch & https://github.com/VCL3D/SphericalViewSynthesis \\ \hline
		2020 & Guizilini \cite{guizilini20203d} & U & PyTorch & https://github.com/TRI-ML/packnet-sfm \\ \hline
		2020 & Klingner \cite{klingner2020self} & U & PyTorch & https://github.com/ifnspaml/SGDepth \\ \hline 		 
		2020 & Peng et al. \cite{peng2019edge} & U & TensorFlow & https://github.com/kspeng/lw-eg-monodepth \\ \hline
		2020 & Shu et al. \cite{shu2020feature} & U & PyTorch & https://github.com/sconlyshootery/FeatDepth \\ \hline		
		2020 & Xue et al. \cite{xue2020toward} & U & PyTorch & https://github.com/TJ-IPLab/DNet \\ \hline \hline
		2017 & Kuznietsov \cite{kuznietsov2017semi} & Semi & TensorFlow & https://github.com/Yevkuzn/semodepth \\ \hline
		2018 & Ramirez \cite{ramirez2018geometry} & Semi & TensorFlow & https://github.com/CVLAB-Unibo/Semantic-Mono-Depth \\ \hline 		
		2019 & Amiri \cite{amiri2019semi} & Semi & TensorFlow & https://github.com/jahaniam/semiDepth \\ \hline \hline
		2018 & Atapour et al. \cite{atapour2018real} & D & Pytorch & https://github.com/atapour/monocularDepth-Inference \\ \hline
		2018 & Guo et al. \cite{guo2018learning} & D & Pytorch & https://github.com/xy-guo/Learning-Monocular-Depth-by-Stereo \\ \hline
		2018 & Zheng et al. \cite{zheng2018t2net} & D & Pytorch & https://github.com/lyndonzheng/Synthetic2Realistic \\ \hline
		2019 & Zhao et al. \cite{zhao2019geometry} & D & Pytorch &  https://github.com/sshan-zhao/GASDA \\ \hline \hline		
	\end{tabular}
\end{table*}
\begin{table*}[!tb]
	\caption{Comparison of many monocular depth estimation methods on the KITTI dataset \cite{geiger2013vision} using the data split in \cite{eigen2014depth}. Depth range from 0m to 80m. ``T": traditional, ``S": supervised, ``U": unsupervised, ``Semi": semi-supervised, ``D": domain adaptation, ``GPU": running-time (\textbf{ms}) tested on for a single forward pass and ``-": not available. The best result of supervised learning-based method is shown in {\color{red}{red}} and {\color{red}{\textbf{bold}}} values, the best result of unsupervised learning-based method is shown in {\color{blue}{blue}} and {\color{blue}{\textbf{bold}}} values, the best result of semi-supervised learning-based method is shown in {\color{green}{green}} and {\color{green}{\textbf{bold}}} values, and the best result of domain adaptation method is shown in {\color{cyan}{cyan}} and {\color{cyan}{\textbf{bold}}} values.}
	\renewcommand{\arraystretch}{1.3}
	\label{table: comparison_of_mde_on_kitti}
	\centering
	\begin{tabular}{c | c | c | c | c | c | c | c | c | c | c | c} \hline\hline
		Year & Algorithm & Type & Abs Rel & Sq Rel & RMSE & RMSE log & {${\delta_{1}}$} & {${\delta_{2}}$} & {${\delta_{3}}$} & GPU & Device \\ \hline \hline
		2008 & Saxena et al. \cite{saxena2008make3d}  & T & 0.412 & 5.712 & 9.635 & 0.444 & 0.556 & 0.752 & 0.870 & - & - \\ \hline \hline
		2014 & Eigen et al. \cite{eigen2014depth} & S & 0.190 & 1.515 & 7.156 & 0.270 & 0.692 & 0.899 & 0.967 & 13 & NVidia Titan Black \\ \hline
		2017 & Cao et al. \cite{cao2017estimating} & S & 0.115 & - & 4.712 & 0.198 & 0.887 & 0.963 & 0.982 & - & - \\ \hline
		2017 & Kuznietsov et al. \cite{kuznietsov2017semi} & S & 0.122 & 0.763 & 4.815 & 0.194 & 0.845 & 0.957 & 0.987 & 48 & Nvidia GTX 980Ti \\ \hline
		2018 & Alhashim and Wonka \cite{alhashim2018high} & S & 0.093 & 0.589 & 4.170 & 0.171 & 0.886 & 0.965 & 0.986 & 333.3 & Jetson AGX Xavier \\ \hline		
		2018 & Fu et al. \cite{fu2018deep} & S & 0.072 & 0.307 & 2.727 & 0.120 & 0.932 & 0.984 & 0.994 & 500 & - \\ \hline
		2018 & Guo et al. \cite{guo2018learning} & S & 0.105 & 0.717 & 4.422 & 0.183 & 0.874 & 0.959 & 0.983 & - & - \\ \hline		
		2018 & Gurram et al. \cite{gurram2018monocular} & S & 0.100 & 0.601 & 4.298 & 0.174 & 0.874 & 0.966 & 0.989 & - & - \\ \hline
		2018 & Kumar et al. \cite{cs2018depthnet} & S & 0.137 & 1.019 & 5.187 & 0.218 & 0.809 & 0.928 & 0.971 & - & - \\ \hline
		2018 & Li et al. \cite{li2018monocular} & S & 0.104 & 0.697 & 4.513 & 0.164 & 0.868 & 0.967 & 0.990 & - & - \\ \hline
		2019 & Lee et al. \cite{lee2019big} & S & 0.059 & 0.241 & 2.756 & 0.096 & 0.956 & 0.993 & 0.998 & - & - \\ \hline  
		2019 & Wang et al. \cite{wang2019recurrent} & S & 0.088 & 0.245 & {\color{red}{\textbf{1.949}}} & 0.127 & 0.915 & 0.984 & 0.996 & - & - \\ \hline
		2019 & Yin et al. \cite{yin2019enforcing} & S & 0.072 & - & 3.258 & 0.117 & 0.938 & 0.990 & 0.998 & - & - \\ \hline
		2020 & Patil et al. \cite{patil2020don} & S & 0.102 & 0.655 & 4.148 & 0.172 & 0.884 & 0.966 & 0.987 & {\color{red}{\textbf{10}}} & - \\ \hline 
		2020 & Wang et al. \cite{wang2020depthnet} & S & 0.103 & 0.511 & 3.916 & - & 0.894 & 0.978 & 0.994 & 71.84 & Jetson AGX Xavier \\ \hline 
		2021 & Bhat et al. \cite{bhat2021adabins} & S & {\color{red}{\textbf{0.058}}} & {\color{red}{\textbf{0.190}}} & 2.360 & {\color{red}{\textbf{0.088}}} & {\color{red}{\textbf{0.964}}} & {\color{red}{\textbf{0.995}}} & {\color{red}{\textbf{0.999}}} & - & - \\ \hline \hline
		2017 & Godard et al. \cite{godard2017unsupervised} & U & 0.148 & 1.344 & 5.927 & 0.247 & 0.862 & 0.960 & 0.964 & 35 & Nvidia Titan-X \\ \hline
		2017 & Kuznietsov et al. \cite{kuznietsov2017semi} & U & 0.308 & 9.367 & 8.700 & 0.367 & 0.752 & 0.904 & 0.952 & 48 & Nvidia GTX 980Ti \\ \hline		
		2017 & Zhou et al. \cite{zhou2017unsupervised} & U & 0.208 & 1.768 & 6.865 & 0.283 & 0.678 & 0.885 & 0.957 & 30 & Nvidia Titan-X \\ \hline			
		2018 & Aleotti et al. \cite{aleotti2018generative} & U & 0.118 & 0.908 & 4.978 & {\color{blue}{\textbf{0.150}}} & 0.855 & 0.948 & 0.976 & - & - \\ \hline
		2018 & Mahjourian et al. \cite{mahjourian2018unsupervised} & U & 0.163 & 1.240 & 6.220 & 0.250 & 0.762 & 0.916 & 0.968 & {\color{blue}{\textbf{10.5}}} & Nvidia GTX 1080 \\ \hline
		2018 & Pilzer et al. \cite{pilzer2018unsupervised} & U & 0.152 & 1.388 & 6.016 & 0.247 & 0.789 & 0.918 & 0.965 & 140 & Nvidia k80 \\ \hline
		2018 & Poggi et al. \cite{poggi2018towards} & U & 0.153 & 1.363 & 6.030 & 0.252 & 0.789 & 0.918 & 0.963 & 20 & Nvidia TiTan-X \\ \hline 	
		2018 & Qi et al. \cite{qi2018geonet} & U & 0.155 & 1.296 & 5.857 & 0.233 & 0.793 & 0.931 & 0.973 & 870 & Nvidia TiTan-X \\ \hline	
		2018 & Zou et al. \cite{zou2018df} & U & 0.150 & 1.124 & 5.507 & 0.223 & 0.806 & 0.933 & 0.973 & - & - \\ \hline		
		2019 & Almalioglu et al. \cite{almalioglu2019ganvo} & U & 0.150 & 1.141 & 5.448 & 0.216 & 0.808 & 0.939 & 0.975 & - & - \\ \hline
		2019 & Bian et al. \cite{bian2019unsupervised} & U & 0.137 & 1.089 & 5.439 & 0.217 & 0.830 & 0.942 & 0.975 & - & - \\ \hline		
		2019 & Godard et al. \cite{godard2019digging} & U & 0.115 & 0.882 & 4.701 & 0.190 & 0.879 & 0.961 & 0.982 & - & - \\ \hline	 		
		2019 & Ranjan et al. \cite{ranjan2019competitive} & U & 0.140 & 1.070 & 5.326 & 0.217 & 0.826 & 0.941 & 0.975 & - & - \\ \hline
		2019 & Tosi et al. \cite{tosi2019learning} & U & 0.111 & 0.867 & 4.714 & 0.199 & 0.864 & 0.954 & 0.979 & 160 & Nvidia TiTan-Xp \\ \hline
		2020 & Guizilini et al. \cite{guizilini2020semantically} & U & {\color{blue}{\textbf{0.102}}} & {\color{blue}{\textbf{0.698}}} & {\color{blue}{\textbf{4.381}}} & 0.178 & {\color{blue}{\textbf{0.896}}} & {\color{blue}{\textbf{0.964}}} & {\color{blue}{\textbf{0.984}}} & - & - \\ \hline	
		2020 & Liu et al. \cite{liu2020mininet} & U & 0.141 & 1.080 & 5.264 & 0.216 & 0.825 & 0.941 & 0.976 & 18.57 & Nvidia GTX 1080Ti \\ \hline
		2020 & Zhao et al. \cite{zhao2020towards} & U & 0.113 & 0.704 & 4.581 & 0.184 & 0.871 & 0.961 & {\color{blue}{\textbf{0.984}}} & - & - \\ \hline \hline 
		2017 & Cho et al. \cite{cho2019large} & Semi & 0.099 & 0.748 & 4.599 & 0.183 & 0.880 & 0.959 & 0.983 & - & - \\ \hline
		2017 & Kuznietsov et al. \cite{kuznietsov2017semi} & Semi & 0.113 & 0.741 & 4.621 & 0.189 & 0.862 & 0.960 & 0.986 & 48 & Nvidia GTX 980Ti \\ \hline
		2019 & Amiri et al. \cite{amiri2019semi} & Semi & 0.096 & 0.552 & 3.995 & 0.152 & 0.892 & {\color{green}{\textbf{0.972}}} & {\color{green}{\textbf{0.992}}} & - & - \\ \hline
		2019 & Dos et al. \cite{dos2019sparse} & Semi & 0.123 & 0.641 & 4.524 & 0.199 & 0.881 & 0.966 & 0.986 & - & - \\ \hline
		2020 & Guizilini et al. \cite{guizilini2020robust} & Semi & {\color{green}{\textbf{0.072}}} & {\color{green}{\textbf{0.340}}} & {\color{green}{\textbf{3.265}}} & {\color{green}{\textbf{0.116}}} & {\color{green}{\textbf{0.934}}} & - & - & - & - \\ \hline 
		2020 & Zhao et al. \cite{zhao2020domain} & Semi & 0.143 & 0.927 & 4.679 & 0.246 & 0.798 & 0.922 & 0.968 & - & - \\ \hline \hline
		2018 & Atapouret al. \cite{atapour2018real} & D & 0.110 & 0.929 & 4.726 & 0.194 & {\color{cyan}{\textbf{0.923}}} & {\color{cyan}{\textbf{0.967}}} & 0.984 & {\color{cyan}{\textbf{22.7}}} & Nvidia GTX 1080Ti \\ \hline 
		2018 & Guo et al. \cite{guo2018learning} & D & {\color{cyan}{\textbf{0.096}}} & {\color{cyan}{\textbf{0.641}}} & {\color{cyan}{\textbf{4.095}}} & {\color{cyan}{\textbf{0.168}}} & 0.892 & {\color{cyan}{\textbf{0.967}}} & {\color{cyan}{\textbf{0.986}}} & - & - \\ \hline 
		2019 & Zhao et al. \cite{zhao2019geometry} & D & 0.149 & 1.003 & 4.995 & 0.227 & 0.824 & 0.941 & 0.973 & - & - \\ \hline \hline		 				
	\end{tabular}
	\justify The results of Saxena et al. \cite{saxena2008make3d} are reproduce from Eigen et al. \cite{eigen2014depth}; the running-time of Fu et al. \cite{fu2018deep} is reported in Patil et al. \cite{patil2020don}; the running-time of Alhashim et al. \cite{alhashim2018high} is reported in Wang et al. \cite{wang2020depthnet}.
\end{table*}
\subsection{Depth Estimation with Semi-supervised Learning}
Unsupervised learning methods eliminate the dependence on ground-truth, which is time-consuming and expensive to obtain. However, their accuracy is limited by stereo construction. With this motivation, semi-supervised methods \cite{kuznietsov2017semi, ramirez2018geometry, cho2019large, tian2019semi, amiri2019semi, ji2019semi, guizilini2020robust} use a small amount of labeled data and a large amount of unlabeled data to improve the accuracy of depth estimation. \par 
The general semi-supervised monocular depth estimation (MDE) network works as follows (see Fig. \ref{fig:schemes}(c)). First, the model is trained with a small amount of labeled training data until it achieves good performance. Then the trained network is used with unlabeled training data to produce outputs known as pseudo labels which may not be quite accurate. The labels and input images from the labeled training data are linked with the generated pseudo labels and input images in the unlabeled training data. Finally, the model is trained in the same way as the first step.
\subsubsection{General Semi-supervised Methods} Kuznietsov et al. \cite{kuznietsov2017semi} design the first semi-supervised learning MDE network by combining supervised and unsupervised loss terms together. The network is trained with the image-sparse depth pairs and unlabeled stereo images. The unsupervised learning-based on direct image alignment between the stereo images is utilized to complement supervised training. Experiments prove that the semi-supervised results outperform the supervised and unsupervised results (see Table \ref{table: comparison_of_mde_on_kitti} for numerical indicators). Amiri et al. \cite{amiri2019semi} extend \cite{godard2017unsupervised} to a semi-supervised network by adding sparse ground-truth data as additional labels for supervised learning. In the training stage, LiDAR data is used as the supervisory signal, and rectified stereo images are used for unsupervised training. \par 
Ji et al. \cite{ji2019semi} introduce a semi-supervised adversarial learning network that is trained on a small number of image-depth pairs and a large number of unlabeled monocular images. The proposed framework consists of a generator network for depth estimation and two discriminator networks to measure the quality of the estimated depth map. During training, unlabeled images are passed to the generator to output depth maps. The two discriminator networks provide feedback to the generator as a unified loss to enable the generator output depth map that accords with the natural depth value distribution. Meanwhile, Guizilini et al. \cite{guizilini2020robust} propose a novel supervised loss which optimizes the re-projected depth in the image space. The designed loss term operates under the same conditions as the photometric loss, by re-projecting depth errors back onto the image space. Hence, the depth labels are incorporated into an appearance-based unsupervised learning method and generates a semi-supervised approach. \par 
\subsubsection{Semi-supervised Methods with Other Tasks} Ramirez et al. \cite{ramirez2018geometry} propose a semi-supervised network for joint learning of semantic segmentation and depth estimation. The network has a shared encoder, a depth decoder and a semantic decoder. The depth estimation task is trained with unsupervised image re-projection loss, while semantic segmentation is trained in the supervised manner. During training time, the semantic segmentation branch provides feedback to the encoder which enables a shared feature representation of both tasks. In addition, a cross-domain discontinuity is proposed to improve the performance of depth estimation. Yue et al. \cite{yuesemi} present a semi-supervised depth estimation framework which consists of a symmetric depth estimation network and a pose estimation network. The RGB image and its semantic map are passed to each sub-network of the depth network to produce an initial depth map and a semantic weight map separately. The two are integrated to generate the final depth map. The pose estimation network outputs a 6 DoF pose for view synthesis. The depth estimation network is trained by minimizing the difference between the synthesized view and target view. \par 
Tian and Li \cite{tian2019semi} introduce a confidence learning-based semi-supervised algorithm by stacking a depth network and a confidence network. The depth network can be any MDE network, e.g., \cite{eigen2014depth, laina2016deeper}, which takes an RGB image as input and produces a depth map, while the confidence network incorporates an RGB image image and the produced depth map to generate a spatial confidence map. The produced confidence map is then utilized as the supervisory signal to guide the training of depth network on unlabeled data. Inspired by the student-teacher strategy, Cho et al. \cite{cho2019large} design a semi-supervised learning framework that stacks a stereo matching network and an MDE network. The deep stereo matching network \cite{pang2017cascade} which trained with ground-truth is used as teacher to produce depth maps from the stereo image pairs. Then stereo confidence maps are predicted to cope with the estimation error from the deep stereo network. The generated depth maps and stereo confidence maps are used as ``pseudo ground-truth" to supervise the training of a shallow MDE network. With this method, the MDE network performs as accurately as the deeper teacher network, and yields better performance than directly learning with ground-truth data. \par
Semi-supervised learning methods learn depth from a small amount of labeled data and a large amount of unlabeled data. The labeled data can be some auxiliary information, e.g., sparse depth or semantic maps, which enables the estimated depth maps more accuracy than unsupervised learning methods. Semi-supervised learning methods alleviate the dependence on ground-truth to some extent, however, it still requires a large amount of unlabeled data in training.
\subsection{Monocular Depth Estimation with Domain Adaptation} 
The first three subsections review DNN-based monocular depth estimation (MDE) methods trained on data collected in real world. Recent advances in computer graphics and modern high-level generic graphic platforms such as game engines make it possible to generate a large set of synthetic 3D scenes. With the constructed scenes, researchers can capture a large amount of synthetic images and their corresponding depth maps to train the MDE model. While training MDE models on synthetic data mitigates the cost of collecting real datasets consist of a large set of image-depth pairs, the produced models normally do not generalize well to the real scenes because of the inherent domain gap\footnote{Due to the distinctions in the intrinsic nature of different domains, the model trained on data from one domain is often incapable of performing well on data from another domain.}. In order to tackle this problem, domain adaptation-based methods first train MDE networks on synthetic data to mitigate the effect of domain gap, making the synthetic data representative of real data (see Fig.\ref{fig:schemes}(d)) have been proposed. \par
\subsubsection{Domain Adaptation via Fine-tuning} 
Approaches reviewed in this subsection first train a network on images from a certain domain such as synthetic data, and then fine-tune it on images from the target domain. According to our investigation, DispNet \cite{mayer2016large} is the first work that apply fine-tuning to overcome domain gap for depth estimation. DispNet is first trained on a large synthetic dataset, and then fine-tuned on a smaller dataset with ground-truth. Guo et al. \cite{guo2018learning} first apply synthetic data to train a stereo matching network. Subsequently, the stereo matching network is fine-tuned on real data. Finally, the produced disparity maps from the stereo network are used as ground-truth to train the MDE network. Experimental results demonstrate that \cite{guo2018learning} outperforms Eigen et al. Fine \cite{eigen2014depth}, \cite{godard2017unsupervised, zhou2017unsupervised}, and Kuznietsov et al. supervised \cite{kuznietsov2017semi}. \par
The fine-tuning based methods normally require a certain amount of ground-truth depth from the target domain. However, suitable ground-truth depth is only available for a few benchmark datasets, e.g., KITTI. Furthermore, in practical settings collecting RGB images with corresponding ground-truth depth maps requires expensive sensors (e.g., LIDAR) and accurate calibration. Since this procedure is complicated and costly, collecting enough real data to pursue fine-tuning in the target domain is seldom feasible \cite{tonioni2019unsupervised}. \par
\subsubsection{Domain Adaptation via Data Transformation}
Methods reviewed in this part transform data in one domain to look similar in style to the data from another domain. Atapour-Abarghouei et al. \cite{atapour2018real} introduce a GAN-based style transfer approach to adapt the real data to fit into the distribution approximated by the generator in the depth estimation model. In order to infer depth maps, a stereo matching network is applied to compute disparity from pixel-wise matching. Compared with methods which directly learn from synthetic data, \cite{atapour2018real} generalizes better from synthetic domain to real domain. Zheng et al. \cite{zheng2018t2net} develop an end-to-end trainable framework that consists of an image translation network ($G_{S \to R}$, where $S$ means synthetic and $R$ means real) and a task MDE network ($f_{T}$). The translation network takes as input the synthetic and real training images. For the real images, $G_{S \to R}$ behaves as an autoencoder and uses a reconstruction loss to apply minimal change to the images. For the synthetic data, $G_{S \to R}$ uses a GAN loss to translates synthetic images into the real domain. The translated images are fed to $f_{T}$ to estimate depth maps which are compared to the synthetic ground-truth depth maps. \par 
However, \cite{atapour2018real, zheng2018t2net} does not consider the geometric structure of the natural images from the target domain. Zhao et al. \cite{zhao2019geometry} exploit the epipolar geometry between the stereo images and design a geometry-aware symmetric domain adaptation network (GASDA) for MDE. The designed framework consists of a style transfer network and a depth estimation network. Since the style transfer network considers both real-to-synthetic and synthetic-to-real translations, two depth estimators can be trained on the original synthetic data and the generated realistic data in supervised manners respectively. \par 
The data transformation-based methods achieve domain invariance in terms of visual appearance by mitigating the cross-domain discrepancy in image layout and structure. It suffers a drop in accuracy when dealing with environments that are different in appearance and/or context from the source domain \cite{tonioni2019unsupervised}. Moreover, sudden change of the illumination or the saturation in images may influence the quality of the transformed images, which will impair the performance of depth estimation \cite{jing2019neural}. \par
Domain adaptation enables MDE networks trained on the synthetic data are adapted to real data, which reduces the cost of acquiring ground-truth depth in real-world environments. It is a promising technique for addressing the unavailability of large amounts of labeled real data. \par
\section{Other Related Methods}
In this section, we review methods for constructing a dense depth map on top of a sparse depth map from LIDAR or SLAM. These methods take as input the sparse depth maps and the aligned RGB images to fill-in missing data in the sparse depth maps. The reviewed methods are referred to in the literature as ``depth completion." \par
\subsection{Sparse depth map from LIDAR}
To the best of our knowledge, Liao et al. \cite{liao2017parse} is the first to perform depth completion. Given a partially observed depth map, the first step is to generate a dense reference depth map via projecting the 2D planar depth values along the gravity direction. The reference depth map is concatenated with the corresponding image and passed to the network which combines both classification and regression losses for estimating the continuous depth value. Ma and Karaman \cite{ma2018sparse} concatenate a set of sparse depth points from LIDAR with an RGB image in the channel dimension to train a depth network. Unlike \cite{liao2017parse}, the sparse depth data is randomly sampled from the ground-truth depth image in order to complement the RGB data. \par
Since the depth data and RGB intensities represent different information, Jaritz et al. \cite{jaritz2018sparse} fuse sparse depth data from LIDAR and RGB images in a late fusion method. Specifically, the RGB image and sparse depth are processed by two encoders separately. The generated feature maps are concatenated along the channel axis and then fed to the decoder network to generate a dense depth map. \cite{liao2017parse, ma2018sparse, jaritz2018sparse} use depth data to update model weights in the process of training. Wang et al. \cite{wang2019plug} design a Plug-and-Play (PnP) module to improve the accuracy of existing MDE networks by using sparse depth data in the process of inference. For the general training of the MDE network, the aim is to minimize the error between the estimation $f(I)$ and ground-truth $D^{*}$, with respect to the network $f$ parametrized by $\theta$ through Equation \ref{eq:theta}.
\begin{equation} \label{eq:theta}
	{\theta}^{*} = argmin L(f(I; \theta), D^{*})
\end{equation}
where $L(\cdot, \cdot)$ is the loss function. Both the model parameters $\theta$ and the input $I$ can affect the estimated depth $f(I; \theta)$, but only the parameters are updated in the process of training. The designed PnP module utilizes the gradient computed from the sparse depth map to update the intermediate feature representation, which is a function of $I$. \par
Chen et al. \cite{chen2019learning} design a 2D-3D fusion block for the joint learning of 2D and 3D feature representations. The designed block consists of a multi-scale 2D convolution branch and a 3D continuous convolution branch. These two branches extract features from the RGB image and the sparse depth data separately, and the generated feature maps are fused by element-wise summation. With this design, various sized network can be created by stacking 2D-3D fusion block sequentially. Qiu et al. \cite{qiu2019deeplidar} infer dense depth maps from the sparse depth maps and the RGB images while using surface normals as the intermediate representation. The designed network consists of a color branch and a surface normal branch. These two branches take as input the RGB image and sparse depth respectively. The color branch directly outputs a dense depth map. The surface normal branch first produces a surface normal image which is fused with the sparse input and a confidence mask from the color branch to produce a dense depth map. Depth maps from different branches are then fused by an attention mechanism to compute the final depth. \par    
\subsection{Sparse depth map from SLAM}
Yang et al. \cite{yang2019fast} utilize sparse depth maps from ORB-SLAM to guide the learning of a dense depth map and a confidence map. The RGB image and sparse depth map are separately processed by a convolutional layer and a max pooling layer. The generated feature maps from the RGB image and the sparse depth map are then concatenated together and processed by another convolutional layer. The fused feature maps are passed to an encoder-decoder network to generate the output. Sartipi et al. \cite{sartipi2020deep} use RGB images, learned surface normals and sparse depth from visual-inertial SLAM (VI-SLAM) to infer dense depth maps. Since the depth map from VI-SLAM is more sparse, a sparse-depth enrichment step is performed to increase its density. The enriched sparse depth maps along with the RGB images and surface normals are passed to the depth completion network to produce dense depth maps. \par
Depth completion methods integrate the RGB images and sparse depth information to generate dense depth maps. The RGB images provide color, texture, contextual and scene structure information, while the sparse depth maps provide a rough geometric structure of the scene. The two input data are complementary, thus, this category of methods demonstrate better accuracy rate than MDE methods \cite{liao2017parse, ma2018sparse}. \par
\section{Discussion and Comparison}
In order to evaluate and compare the monocular depth estimation (MDE) methods, we summarize the quantitative results of 42 representative methods on the KITTI dataset \cite{geiger2013vision}. The performance comparison of the summarized methods is listed in Table \ref{table: comparison_of_mde_on_kitti}, including error metric (Abs Rel, Sq Rel, RMSE and RMSE log, \textsl{lower is better}), accuracy metrics ($\delta_{1}$, $\delta_{2}$ and $\delta_{3}$, \textsl{higher is better}) and running-time (GPU). The results listed in Table \ref{table: comparison_of_mde_on_kitti} are from their respective papers. \par
\subsection{Accuracy}
According to Table \ref{table: comparison_of_mde_on_kitti}, all deep learning-based methods show much better results than the traditional Make3D method \cite{saxena2008make3d}. Thus, Make3D is not applicable in any recent application. We observe that the overall development trend of MDE is to push the increase of accuracy. Among the four categories of methods, the supervised learning method generates the best error and accuracy metric results, followed by the semi-supervised, domain adaptation and unsupervised methods. It demonstrates that supervised learning method can learn more representative features from the ground-truth depth. Note that Bhat et al. \cite{bhat2021adabins} generate the best performance among supervised learning methods, suggesting that explicitly utilizing global information at a high resolution decisively improves the performance of MDE. \par 
Regarding the domain adaptation methods \cite{guo2018learning, atapour2018real, zhao2019geometry}, Guo et al. \cite{guo2018learning} yields the best performance. Unlike \cite{atapour2018real} and \cite{zhao2019geometry}, \cite{guo2018learning} first pre-trained with synthetic data and then fine-tuned on real data. It demonstrates that when the fine-tuning dataset is similar to the test dataset, the fine-tuning method performs better than the data transformation method. The best domain adaptation method \cite{guo2018learning} has superior performance to the best unsupervised method \cite{guizilini2020semantically}. Regarding the best semi-supervised method \cite{guizilini2020semantically} and the best domain adaptation method \cite{guo2018learning}, \cite{guizilini2020semantically} outperforms \cite{guo2018learning}. In particular, their accuracy metrics are almost equal, while the error metrics especially Sq Rel, RMSE and RMSE log are highly variable. This suggests that even small amounts of labeled data can make a great contribution to the performance of depth networks. \par 
\subsection{Computational Time}
In Table \ref{table: comparison_of_mde_on_kitti}, we do not show the running-time of all summarized methods because many publications do not report it. Since some authors did not provide enough information to replicate their results, it is impractical to test the running-time on our computer. However, as the number of network parameters affects memory footprint and running-time required to infer depth, we use this information as an additional information to compare the running-time. For example, Poggi et al. \cite{poggi2018towards} has 1.9M parameters and requires 20ms to infer a depth map on a popular Nvidia Titan-X GPU. \par

According to \cite{bhat2021adabins}, the proposed network has 78M parameters, which is 40 times more than Poggi et al. \cite{poggi2018towards}. Therefore, \cite{bhat2021adabins} requires much more time to infer a depth map it is hard to run at a real-time speed on a single GPU. This demonstrates that \cite{bhat2021adabins} can only be applied to accuracy-first tasks. The supervised method by Wang et al. \cite{wang2020depthnet} runs at about 14fps on a Jetson AGX Xavier embedded device, which is close to a real-time speed. Moreover, \cite{wang2020depthnet} has less parameters than \cite{poggi2018towards} (1.75M vs 1.9M) and yields much better results. This suggests that \cite{wang2020depthnet} is suitable for real-time tasks. Among unsupervised methods, \cite{mahjourian2018unsupervised, poggi2018towards, liu2020mininet} show the three fastest speeds. Although the running-time is tested on different GPUs, the order of GPU computation capacity is ``Titan-X $>$ 1080Ti $>$ 1080". Regarding the error and accuracy metrics, \cite{liu2020mininet} is inferior to \cite{mahjourian2018unsupervised} and \cite{poggi2018towards}. In addition, \cite{liu2020mininet} runs faster than \cite{poggi2018towards} on a less powerful GPU. Therefore, Liu et al. \cite{liu2020mininet} can be applied to real-time tasks where large amount of labeled data is not available. \par
\section{Applications in Robotics}
Autonomous vehicles need to detect obstacles, other cars and pedestrians and depth estimation is a fundamental component required to do this in a 3D environment. Depth estimation is a basic component in perceiving the 3D environment. Although autonomous vehicles and robots can perceive depth information through LIDAR, they only produce sparse depth maps. The sparsity of these depth measurements makes it hard to meet the perception requirements needed for safe self-driving car applications. Monocular depth estimation (MDE) estimates dense depth maps from single images. The resulting dense depth maps have the potential to provide the absolute distances to surfaces of objects in real-time with a single sensor, whilst meeting the requirements of autonomous navigation and obstacle avoidance systems \cite{yurtsever2020survey}. \par 
Autonomous vehicles operate in real-world environments where real-time performance is crucial. Moreover, small autonomous vehicle platforms (e.g., micro aerial vehicles or mini ground vehicles) normally have limited memory and computational resource. The onboard sensor on such platforms may be limited to a monocular RGB camera, and no additional information (e.g., sparse depth point clouds) may be present. Motivated by this fact, lightweight CNNs \cite{howard2017mobilenets, sandler2018mobilenetv2, romera2017erfnet} and monocular depth estimation networks \cite{spek2018cream, wofk2019fastdepth, poggi2018towards, liu2020mininet, oh2020rrnet} that run on real-time embedded devices have been developed on top of depthwise separable convolutions, factorized convolutions or network architecture search techniques. It is worth noting that these lightweight implementations \cite{spek2018cream, wofk2019fastdepth, poggi2018towards, liu2020mininet, oh2020rrnet} achieve a real-time speed on mobile platforms (e.g., Nvidia-TX2 or Jetson AGX Xavier GPU) and produce more accurate depth maps than their traditional counterparts (e.g., \cite{saxena2008make3d, karsch2014depth}) which cannot run in real-time. Therefore, it enables the autonomous vehicles or robots to perceive more accurate depth information than \cite{saxena2008make3d, karsch2014depth}, whilst not limiting the reactive speed of these vehicles. \par
Due to the relatively low cost, size and energy consumption, MDE has been applied to the task of ego-motion estimation \cite{yin2017scale, luo2018real, spek2018cream, yang2018deep, mendes2020deep}, obstacle avoidance \cite{michels2005high, alvarez2016collision, chakravarty2017cnn, yang2019fast, zhang2019monocular, lin2020robust} and scene understanding \cite{scharwachter2015low, jiang2018self, rojas2018landingzone}. \par
\subsection{Ego-motion Estimation}
Systems for calculating ego-motion from vision generally need an absolute range sensor to provide scale to the visual motions. MDE can provide the range measurements needed to provide this information. However, in some cases this can be done using inertial sensing with an appropriate sensor fusion filter. Li et al. \cite{li2016metric} use feature depth and onboard Inertial Measurement Unit (IMU) data to compute optic flow to estimate the motion of UAV. Such techniques need the vehicle to be constantly moving and will fail if there is not motion, and are prone to noise. Wang et al. \cite{wang2015real} design an ego-motion estimation method for UAV by fusing data from an RGB-D camera and an IMU. The utilized depth camera suffers from a limited measurement range (0.5m-4m), which can be replaced by a real-time MDE algorithm, such as \cite{spek2018cream, wofk2019fastdepth, poggi2018towards, liu2020mininet, oh2020rrnet}. \par
DNNs can predict the absolute scale information in the process of MDE, it is helpful in tackling the scale ambiguity and drift problem in monocular ego-motion estimation and improving the mapping process. \cite{yin2017scale, yang2018deep, mendes2020deep} incorporate CNN-based depth estimations into monocular visual odometry (VO). The obtained VO algorithms show robustness to scale drift and achieve comparable performance to stereo VO methods. Tateno et al. \cite{tateno2017cnn} fuse CNN estimated dense depth maps with semi-dense depth measurements from SLAM \cite{engel2014lsd} to solve the scale ambiguity and drift problem of monocular SLAM. In order to improve the computing speed, dense depth maps only computed from every key-frame. In addition, LOO et al. \cite{loo2019cnn} combine the semi-direct visual odometry (SVO) with a depth estimation CNN \cite{godard2017unsupervised}. The added depth estimation network provides depth priors in the map points initialization process when a key-frame is selected. With the prior knowledge of the scene geometry, the proposed CNN-SVO is able to obtain a much better estimate of the mean and a smaller initial variance of the depth-filter than the original SVO. \par
The fusion of depth maps and VO or V-SLAM algorithms improves the performance of ego-motion estimation \cite{tateno2017cnn, loo2019cnn}, while the applied depth estimation CNNs \cite{laina2016deeper, godard2017unsupervised} require a high-end GPU (e.g., Titan X) to achieve a real-time speed. This limits the application of \cite{tateno2017cnn, loo2019cnn} in small sized platforms with limited memory and computational resource. Spek et al. \cite{spek2018cream} integrate the estimated depth maps from a lightweight CNN with ORB-SLAM2 system \cite{mur2017orb}. The fused system runs tracking and mapping on mobile platforms at a real-time speed while effectively reducing scale-drift and improving the accuracy of a standard monocular SLAM system. \par
\subsection{Obstacle Avoidance}
Depth maps contain information about the distance between the surface of objects to the camera \cite{matthies2014stereo, brockers2016vision}. With the estimated depth maps, it is possible for autonomous vehicles or robots to perceive the environment and achieve the goal of avoiding obstacles in stationary scenes \cite{michels2005high, chakravarty2017cnn, zhang2019monocular, alvarez2016collision, yang2019fast}. Michels et al. \cite{michels2005high} use a supervised learning algorithm to learn depth cues that can accurately represent the distance of the nearest obstacles in the scene. The estimated depth information is then converted to steering commands for controlling a ground vehicle in the static outdoor environments. Chakravarty et al. \cite{chakravarty2017cnn} pass depth maps to a behaviour arbitration-based control algorithm to guide a UAV which flies at a particular height in indoor environment to avoid obstacles. Given a depth map, a vertical and horizontal strip through the middle of the depth map is selected. Then the averaged depth values within each vertical and horizontal bin $vert_i$ and $horz_i$ are used as depth values $d_i$ from the angle. These depth values $d_i$ are used to compute an angular velocity for steering the UAV away from obstacles. Zhang et al. \cite{zhang2019monocular} use the estimated depth maps to compute the rotation angle to guide a UAV to avoid obstacles and fly towards a destination. \par
Depth maps can also be used to select collision-free waypoints in order to steer robots in a safe path. Alvarez et al. \cite{alvarez2016collision} first compute a dense depth map from a small set of consecutive images. Then the depth map is applied to generate the next obstacle-free waypoints to proceed in a forward direction. To this end, the most distant point in 3D space reachable by the UAV without collisions will be computed. Yang et al. \cite{yang2019fast} design an Ego Dynamic Space (EDS)-based obstacle avoidance method by embedding the dynamic motion constrains of the UAV and the confidence values into the spatial depth map. With the estimated depth and confidence maps, the distances $D_{eff}$ to the obstacles can be written as in Equation {\ref{eq:d_eff}}. 
\begin{equation}\label{eq:d_eff}
\begin{aligned}
D_{eff} & = D - D_{brake} - D_{error} \\
& = D - (vT - \frac{aT^{2}}{2}) - D_{error}
\end{aligned}
\end{equation}
where $D$ is the perceived distance to the obstacles (i.e. the estimated depth map), $D_{brake}$ is the deceleration distance to stop a UAV which moves at velocity $v$ using deceleration $a$ in a sampling interval $T$, and $D_{error}$ is the depth measurement error which is computed by $D_{error} = -ln(C)$, \textsl{C} is the estimated confidence. Subsequently, $D_{eff}$ is converted into a binary depth map, which includes collision-free regions that a UAV can move safely. \par
It should be noted that the CNNs used in \cite{chakravarty2017cnn, yang2019fast, zhang2019monocular} were trained on data recorded through cameras on moving ground vehicles, which is limited in viewing angles in recorded images. Moreover, the ground vehicles normally move in constrained environments such as roads or corridors. The onboard cameras only encounter a limited subset of motion types, and do not completely explore the 3D environment. This results in the perspective and angle of view of the captured images being different from the UAV view. Therefore, this may raise concerns about the generalization potential of the trained CNNs to the application of obstacle avoidance for UAVs \cite{fonder2019mid}. \par
Depth information offers opportunities and complexities for the determination of the collision-point and time-to-collision when a robot is navigating in a dynamic environment. In particular, depth information could offer the robot a space to navigate in an otherwise more constrained environment. However, error in depth estimation could have a profound impact on a robot’s ability to estimate a collision-point with a moving object and subsequently, the time-to-collision with that object. While there exists an extensive literature on collision avoidance research in dynamic 3D environments \cite{wojke2012moving, wang2012could, cherubini2014autonomous}, including research on point-clouds from sensors such as LIDAR, literature tackling collision avoidance in dynamic 3D environments with depth estimation is almost non-existent. \par
\subsection{Scene Understanding}
Autonomous vehicles and robots require a full understanding of the geometric structure of environment to interact with it. The task of scene understanding is to obtain 3D geometric information from 2D image. Depth maps encode the 3D structure of the scene, which helps to resolve ambiguities and to avoid a physical implausible labeling \cite{scharwachter2015low}. Scharwachter and Franke \cite{scharwachter2015low} propose an approach to infer the coarse layout of street scenes from color, texture and depth information. \par
When an agent moves through the world, the apparent motion of scene elements is usually inversely proportional to their depth. For example, as the agent moves, faraway mountains do not move much, while nearby trees move a lot. Inspired by this point, Jiang et al. \cite{jiang2018self} use the depth estimation network as the base for city scene understanding. They first train a deep network to infer relative scene depth from single images, and then fine-tune it for tasks such as semantic segmentation, joint semantic reasoning of road segmentation and car detection. \par
Rojas-Perez et al. \cite{rojas2018landingzone} propose a depth estimation-based landing zone detection method for UAVs. The detection is divided into two stages: depth estimation from single aerial images and classification of possible landing zones. Based on the extracted patches from RGB images, a multi-layer CNN architecture is designed and trained to infer depth from aerial images. The obtained depth map is then fed to another CNN to detect possible landing zones for UAV. \par
\section{Conclusion} 
In this paper, we presented the first comprehensive survey of monocular depth estimation (MDE). We reviewed the literature from the structure from motion-based methods, traditional handcrafted feature-based methods to the state-of-the-art deep learning-based methods. We also summarized the publically available datasets, commonly used performance evaluation metrics, and open-source implementations of some representative methods. In addition, we compared and analyzed the performance of 42 representative methods from different perspectives, including error, accuracy and running-time metrics. Since MDE plays an important role in robotics, we provided a review of the application of MDE in ego-motion estimation, obstacle avoidance and scene understanding. Based on the reviewed literature, we concluded that the promising future research directions of MDE may focus on but are not limited to the following aspects: \par 
\textbf{Collecting rich scene datasets}: Deep learning-based models show great performance in MDE. However, training robust models requires a dataset consists of various scenes, in order that the models can learn various scene features. Compared with real-world environments which include complex scenarios such as moving objects, cluttered scenes, occlusions, illumination changes and weather changes, the existing public datasets are not rich enough. Specifically, these datasets focus on certain scenes such as indoor \cite{silberman2012indoor}, driving \cite{geiger2013vision}, campus \cite{saxena2008make3d} and forest \cite{niu2020low}. Thus, it is essential to collect datasets that encompass richness and high-diversity of environmental scenarios. \par
\textbf{Real-time monocular depth estimation with accuracy and efficiency balance}: The overall development trend of MDE is to push the increase of accuracy using extremely deep CNNs or by designing a complex network architecture, which are computationally expensive for current mobile computational devices which have limited memory and computational capability. Therefore, it is difficult for these networks to be deployed in small sized robots which depend on mobile computational devices. Under this context, researchers have begun to develop real-time MDE methods \cite{poggi2018towards, wofk2019fastdepth, liu2020mininet, wang2020depthnet}. However, the accuracy of these methods is inferior to state-of-the-art methods. Therefore, developing real-time MDE network is assumed to achieve the trade-off between accuracy and efficiency. \par
\textbf{Monocular depth estimation with domain adaptation}: The training of a supervised MDE network requires a large amount of ground-truth data. However, collecting these ground-truth data requires LIDAR or RGB-D cameras, which increase the cost. With computer graphic techniques, it is easier to obtain a large set of synthetic images and its corresponding depth maps. Applying domain adaptation techniques to training MDE models on synthetic and transferring it to real data seems to be a popular direction in MDE. \par
\textbf{Semi-supervised monocular depth estimation}: The training of supervised MDE network relies on a large amount of labeled data. The process of collecting depth maps is time-consuming, expensive and inefficient. Unsupervised methods do not need the ground-truth data, while suffering from lower accuracy. Developing semi-supervised methods that are trained on a small number of labeled images and a large number of unlabeled images is of great importance for reducing labor costs and improving prediction accuracy. \par 
\textbf{Depth estimation with information fusion}: Depth estimation with multiple sources/modalities of data, such as sparse depth maps, optical flow and surface normal, show the better performance. Some open questions include: how to apply well-designed depth estimation to different modalities of data, how to efficiently fuse different information to improve the accuracy of depth estimation and how to use optical flow to handle independently moving objects in a dynamic scene? \par
\textbf{Interpretability of monocular depth estimation networks}: While deep learning-based depth estimation has achieved remarkable improvement in accuracy, there remain questions about these networks. For example, what exactly are depth estimation networks learning? What is a minimal network architecture that can achieve a certain accuracy? Although studies in \cite{hu2019visualization, dijk2019neural} explored the mechanism of depth estimation networks, a specific study of the underlying behavior/dynamics of these networks was not available. Thus, the research on the interpretability of depth estimation networks is an important topic. \par
\bibliography{references.bib}

\begin{thebibliography}{100}
\providecommand{\url}[1]{#1}
\csname url@samestyle\endcsname
\providecommand{\newblock}{\relax}
\providecommand{\bibinfo}[2]{#2}
\providecommand{\BIBentrySTDinterwordspacing}{\spaceskip=0pt\relax}
\providecommand{\BIBentryALTinterwordstretchfactor}{4}
\providecommand{\BIBentryALTinterwordspacing}{\spaceskip=\fontdimen2\font plus
\BIBentryALTinterwordstretchfactor\fontdimen3\font minus
  \fontdimen4\font\relax}
\providecommand{\BIBforeignlanguage}[2]{{%
\expandafter\ifx\csname l@#1\endcsname\relax
\typeout{** WARNING: IEEEtran.bst: No hyphenation pattern has been}%
\typeout{** loaded for the language `#1'. Using the pattern for}%
\typeout{** the default language instead.}%
\else
\language=\csname l@#1\endcsname
\fi
#2}}
\providecommand{\BIBdecl}{\relax}
\BIBdecl

\bibitem{tateno2017cnn}
K.~Tateno, F.~Tombari, I.~Laina, and N.~Navab, ``{CNN-SLAM}: Real-time dense
  monocular {SLAM} with learned depth prediction,'' in \emph{CVPR}, 2017, pp.
  6243--6252.

\bibitem{yang2019fast}
X.~Yang, J.~Chen, Y.~Dang, H.~Luo, Y.~Tang, C.~Liao, P.~Chen, and K.-T. Cheng,
  ``Fast depth prediction and obstacle avoidance on a monocular drone using
  probabilistic convolutional neural network,'' \emph{IEEE TITS}, 2019.

\bibitem{jiang2018self}
H.~Jiang, G.~Larsson, M.~Maire Greg~Shakhnarovich, and E.~Learned-Miller,
  ``Self-supervised relative depth learning for urban scene understanding,'' in
  \emph{ECCV}, 2018, pp. 19--35.

\bibitem{forouher2016sensor}
D.~Forouher, M.~G. Besselmann, and E.~Maehle, ``Sensor fusion of depth camera
  and ultrasound data for obstacle detection and robot navigation,'' in
  \emph{ICARCV}, 2016, pp. 1--6.

\bibitem{wang2011global}
L.~Wang and R.~Yang, ``Global stereo matching leveraged by sparse ground
  control points,'' in \emph{CVPR}, 2011, pp. 3033--3040.

\bibitem{muresan2015improving}
M.~P. Muresan, M.~Negru, and S.~Nedevschi, ``Improving local stereo algorithms
  using binary shifted windows, fusion and smoothness constraint,'' in
  \emph{ICCP}, 2015, pp. 179--185.

\bibitem{spangenberg2014large}
R.~Spangenberg, T.~Langner, S.~Adfeldt, and R.~Rojas, ``Large scale semi-global
  matching on the cpu,'' in \emph{IV}, 2014, pp. 195--201.

\bibitem{wedel2006realtime}
A.~Wedel, U.~Franke, J.~Klappstein, T.~Brox, and D.~Cremers, ``Realtime depth
  estimation and obstacle detection from monocular video,'' in \emph{JPRS},
  2006, pp. 475--484.

\bibitem{prakash2014sparse}
C.~D. Prakash, J.~Li, F.~Akhbari, and L.~J. Karam, ``Sparse depth calculation
  using real-time key-point detection and structure from motion for advanced
  driver assist systems,'' in \emph{ISVC}, 2014, pp. 740--751.

\bibitem{ha2016high}
H.~Ha, S.~Im, J.~Park, H.-G. Jeon, and I.~So~Kweon, ``High-quality depth from
  uncalibrated small motion clip,'' in \emph{CVPR}, 2016, pp. 5413--5421.

\bibitem{javidnia2017accurate}
H.~Javidnia and P.~Corcoran, ``Accurate depth map estimation from small
  motions,'' in \emph{ICCV}, 2017, pp. 2453--2461.

\bibitem{torralba2002depth}
A.~Torralba and A.~Oliva, ``Depth estimation from image structure,'' \emph{IEEE
  TPAMI}, vol.~24, no.~9, pp. 1226--1238, 2002.

\bibitem{saxena2006learning}
A.~Saxena, S.~H. Chung, and A.~Y. Ng, ``Learning depth from single monocular
  images,'' in \emph{NIPS}, 2006, pp. 1161--1168.

\bibitem{jung2010depth}
J.-I. Jung and Y.-S. Ho, ``Depth map estimation from single-view image using
  object classification based on bayesian learning,'' in \emph{3DTV
  Conference}, 2010, pp. 1--4.

\bibitem{liu2010single}
B.~Liu, S.~Gould, and D.~Koller, ``Single image depth estimation from predicted
  semantic labels,'' in \emph{CVPR}, 2010, pp. 1253--1260.

\bibitem{ladicky2014pulling}
L.~Ladicky, J.~Shi, and M.~Pollefeys, ``Pulling things out of perspective,'' in
  \emph{CVPR}, 2014, pp. 89--96.

\bibitem{liu2014discrete}
M.~Liu, M.~Salzmann, and X.~He, ``Discrete-continuous depth estimation from a
  single image,'' in \emph{CVPR}, 2014, pp. 716--723.

\bibitem{karsch2014depth}
K.~Karsch, C.~Liu, and S.~B. Kang, ``Depth transfer: Depth extraction from
  video using non-parametric sampling,'' \emph{IEEE TPAMI}, vol.~36, no.~11,
  pp. 2144--2158, 2014.

\bibitem{raza2015depth}
S.~H. Raza, O.~Javed, A.~Das, H.~Sawhney, H.~Cheng, and I.~Essa, ``Depth
  extraction from videos using geometric context and occlusion boundaries,''
  \emph{arXiv preprint arXiv:1510.07317}, 2015.

\bibitem{eigen2014depth}
D.~Eigen, C.~Puhrsch, and R.~Fergus, ``Depth map prediction from a single image
  using a multi-scale deep network,'' in \emph{NIPS}, 2014, pp. 2366--2374.

\bibitem{eigen2015predicting}
D.~Eigen and R.~Fergus, ``Predicting depth, surface normals and semantic labels
  with a common multi-scale convolutional architecture,'' in \emph{ICCV}, 2015,
  pp. 2650--2658.

\bibitem{laina2016deeper}
I.~Laina, C.~Rupprecht, V.~Belagiannis, F.~Tombari, and N.~Navab, ``Deeper
  depth prediction with fully convolutional residual networks,'' in \emph{3DV},
  2016, pp. 239--248.

\bibitem{cao2017estimating}
Y.~Cao, Z.~Wu, and C.~Shen, ``Estimating depth from monocular images as
  classification using deep fully convolutional residual networks,'' \emph{IEEE
  TCSVT}, vol.~28, no.~11, pp. 3174--3182, 2017.

\bibitem{xu2017multi}
D.~Xu, E.~Ricci, W.~Ouyang, X.~Wang, and N.~Sebe, ``Multi-scale continuous
  {CRFs} as sequential deep networks for monocular depth estimation,'' in
  \emph{CVPR}, 2017, pp. 5354--5362.

\bibitem{li2018monocular}
B.~Li, Y.~Dai, and M.~He, ``Monocular depth estimation with hierarchical fusion
  of dilated {CNNs} and soft-weighted-sum inference,'' \emph{PR}, vol.~83, pp.
  328--339, 2018.

\bibitem{hu2019revisiting}
J.~Hu, M.~Ozay, Y.~Zhang, and T.~Okatani, ``Revisiting single image depth
  estimation: Toward higher resolution maps with accurate object boundaries,''
  in \emph{WACV}, 2019, pp. 1043--1051.

\bibitem{wofk2019fastdepth}
D.~Wofk, F.~Ma, T.-J. Yang, S.~Karaman, and V.~Sze, ``{FastDepth}: Fast
  monocular depth estimation on embedded systems,'' in \emph{ICRA}, 2019, pp.
  6101--6108.

\bibitem{chavan2015depth}
A.~Chavan and R.~S. Gudodagi, ``Depth extraction from video: a survey,''
  \emph{IJIRAE}, vol.~2, pp. 199--204, 2015.

\bibitem{mohan2015review}
D.~Mohan and A.~R. Ram, ``A review on depth estimation for computer vision
  applications,'' \emph{Positions}, vol.~4, no.~11, 2015.

\bibitem{jamwal2016survey}
N.~Jamwal, N.~Jindal, and K.~Singh, ``A survey on depth map estimation
  strategies,'' in \emph{ICSP}, 2016.

\bibitem{bahadur2017literature}
S.~Bahadur, R.~Shrestha, Y.~Sumaharshini, G.~R. Teja, and K.~N, ``Literature
  review on various depth estimation methods for an image,'' \emph{IJRG},
  vol.~5, 2017.

\bibitem{bhoi2019monocular}
A.~Bhoi, ``Monocular depth estimation: a survey,'' \emph{arXiv preprint
  arXiv:1901.09402}, 2019.

\bibitem{laga2019survey}
H.~Laga, ``A survey on deep learning architectures for image-based depth
  reconstruction,'' \emph{arXiv preprint arXiv:1906.06113}, 2019.

\bibitem{vyshna2019literature}
R.~Vyshna and S.~Priya, ``Literature review on depth estimation using a single
  image,'' \emph{IJARIIT}, vol.~5, pp. 1760--1763, 2019.

\bibitem{liu2020survey}
Y.~Liu, J.~Jiang, J.~Sun, L.~Bai, and Q.~Wang, ``A survey of depth estimation
  based on computer vision,'' in \emph{DSC}, 2020, pp. 135--141.

\bibitem{saxena2008make3d}
A.~Saxena, M.~Sun, and A.~Y. Ng, ``{Make3D}: Learning {3D} scene structure from
  a single still image,'' \emph{IEEE TPAMI}, vol.~31, no.~5, pp. 824--840,
  2008.

\bibitem{silberman2012indoor}
N.~Silberman, D.~Hoiem, P.~Kohli, and R.~Fergus, ``Indoor segmentation and
  support inference from {RGB-D} images,'' in \emph{ECCV}, 2012, pp. 746--760.

\bibitem{geiger2013vision}
A.~Geiger, P.~Lenz, C.~Stiller, and R.~Urtasun, ``Vision meets robotics: The
  {KITTI} dataset,'' \emph{IJRR}, vol.~32, no.~11, pp. 1231--1237, 2013.

\bibitem{cordts2016cityscapes}
M.~Cordts, M.~Omran, S.~Ramos, T.~Rehfeld, M.~Enzweiler, R.~Benenson,
  U.~Franke, S.~Roth, and B.~Schiele, ``The cityscapes dataset for semantic
  urban scene understanding,'' in \emph{CVPR}, 2016, pp. 3213--3223.

\bibitem{gaidon2016virtual}
A.~Gaidon, Q.~Wang, Y.~Cabon, and E.~Vig, ``Virtual worlds as proxy for
  multi-object tracking analysis,'' in \emph{CVPR}, 2016, pp. 4340--4349.

\bibitem{cabon2020virtual}
Y.~Cabon, N.~Murray, and M.~Humenberger, ``Virtual {KITTI} 2,'' \emph{arXiv
  preprint arXiv:2001.10773}, 2020.

\bibitem{sturm2012benchmark}
J.~Sturm, N.~Engelhard, F.~Endres, W.~Burgard, and D.~Cremers, ``A benchmark
  for the evaluation of {RGB-D SLAM} systems,'' in \emph{IROS}, 2012, pp.
  573--580.

\bibitem{song2015sun}
S.~Song, S.~P. Lichtenberg, and J.~Xiao, ``{SUN RGB-D}: {A} {RGB-D} scene
  understanding benchmark suite,'' in \emph{CVPR}, 2015, pp. 567--576.

\bibitem{chen2016single}
W.~Chen, Z.~Fu, D.~Yang, and J.~Deng, ``Single-image depth perception in the
  wild,'' in \emph{NIPS}, 2016, pp. 730--738.

\bibitem{wasenmuller2016corbs}
O.~Wasenm{\"u}ller, M.~Meyer, and D.~Stricker, ``{CoRBS}: Comprehensive {RGB-D}
  benchmark for {SLAM} using {Kinect v2},'' in \emph{WACV}, 2016, pp. 1--7.

\bibitem{armeni2017joint}
I.~Armeni, S.~Sax, A.~R. Zamir, and S.~Savarese, ``Joint {2D-3D}-semantic data
  for indoor scene understanding,'' \emph{arXiv preprint arXiv:1702.01105},
  2017.

\bibitem{schops2017multi}
T.~Schops, J.~L. Schonberger, S.~Galliani, T.~Sattler, K.~Schindler,
  M.~Pollefeys, and A.~Geiger, ``A multi-view stereo benchmark with
  high-resolution images and multi-camera videos,'' in \emph{CVPR}, 2017, pp.
  3260--3269.

\bibitem{chang2017matterport3d}
A.~Chang, A.~Dai, T.~Funkhouser, M.~Halber, M.~Niessner, M.~Savva, S.~Song,
  A.~Zeng, and Y.~Zhang, ``Matterport3{D}: Learning from {RGB-D} data in indoor
  environments,'' \emph{arXiv preprint arXiv:1709.06158}, 2017.

\bibitem{dai2017scannet}
A.~Dai, A.~X. Chang, M.~Savva, M.~Halber, T.~Funkhouser, and M.~Nie{\ss}ner,
  ``{ScanNet}: Richly-annotated {3D} reconstructions of indoor scenes,'' in
  \emph{CVPR}, 2017, pp. 5828--5839.

\bibitem{mccormac2017scenenet}
J.~McCormac, A.~Handa, S.~Leutenegger, and A.~J. Davison, ``{SceneNet} {RGB-D}:
  Can {5M} synthetic images beat generic {ImageNet} pre-training on indoor
  segmentation?'' in \emph{ICCV}, 2017, pp. 2678--2687.

\bibitem{song2017semantic}
S.~Song, F.~Yu, A.~Zeng, A.~X. Chang, M.~Savva, and T.~Funkhouser, ``Semantic
  scene completion from a single depth image,'' in \emph{CVPR}, 2017, pp.
  1746--1754.

\bibitem{li2018megadepth}
Z.~Li and N.~Snavely, ``{MegaDepth}: Learning single-view depth prediction from
  internet photos,'' in \emph{CVPR}, 2018, pp. 2041--2050.

\bibitem{mancini2018j}
M.~Mancini, G.~Costante, P.~Valigi, and T.~A. Ciarfuglia, ``{J-MOD} 2: Joint
  monocular obstacle detection and depth estimation,'' \emph{IEEE RAL}, vol.~3,
  no.~3, pp. 1490--1497, 2018.

\bibitem{marcu2018safeuav}
A.~Marcu, D.~Costea, V.~Licaret, M.~P{\^\i}rvu, E.~Slusanschi, and
  M.~Leordeanu, ``{SafeUAV}: Learning to estimate depth and safe landing areas
  for {UAVs} from synthetic data,'' in \emph{ECCV Workshops}, 2018, pp. 43--58.

\bibitem{zioulis2018omnidepth}
N.~Zioulis, A.~Karakottas, D.~Zarpalas, and P.~Daras, ``{OmniDepth}: Dense
  depth estimation for indoors spherical panoramas,'' in \emph{ECCV}, 2018, pp.
  448--465.

\bibitem{wu2019depth}
S.~Wu, H.~Zhao, and S.~Sun, ``Depth estimation from infrared video using
  local-feature-flow neural network,'' \emph{IJMLC}, vol.~10, no.~9, pp.
  2563--2572, 2019.

\bibitem{cho2019large}
J.~Cho, D.~Min, Y.~Kim, and K.~Sohn, ``A large {RGB-D} dataset for
  semi-supervised monocular depth estimation,'' \emph{arXiv preprint
  arXiv:1904.10230}, 2019.

\bibitem{yang2019drivingstereo}
G.~Yang, X.~Song, C.~Huang, Z.~Deng, J.~Shi, and B.~Zhou, ``{DrivingStereo}: A
  large-scale dataset for stereo matching in autonomous driving scenarios,'' in
  \emph{CVPR}, 2019, pp. 899--908.

\bibitem{vasiljevic2019diode}
I.~Vasiljevic, N.~Kolkin, S.~Zhang, R.~Luo, H.~Wang, F.~Z. Dai, A.~F. Daniele,
  M.~Mostajabi, S.~Basart, M.~R. Walter \emph{et~al.}, ``{DIODE}: A dense
  indoor and outdoor depth dataset,'' \emph{arXiv preprint arXiv:1908.00463},
  2019.

\bibitem{fonder2019mid}
M.~Fonder and M.~Van~Droogenbroeck, ``{Mid-Air}: A multi-modal dataset for
  extremely low altitude drone flights,'' in \emph{CVPR Workshops}, 2019, pp.
  0--0.

\bibitem{niu2020low}
C.~Niu, D.~Tarapore, and K.-P. Zauner, ``Low-viewpoint forest depth dataset for
  sparse rover swarms,'' \emph{arXiv preprint arXiv:2003.04359}, 2020.

\bibitem{jin2020geometric}
L.~Jin, Y.~Xu, J.~Zheng, J.~Zhang, R.~Tang, S.~Xu, J.~Yu, and S.~Gao,
  ``Geometric structure based and regularized depth estimation from 360 indoor
  imagery,'' in \emph{CVPR}, 2020, pp. 889--898.

\bibitem{liu2015learning}
F.~Liu, C.~Shen, G.~Lin, and I.~Reid, ``Learning depth from single monocular
  images using deep convolutional neural fields,'' \emph{IEEE TPAMI}, vol.~38,
  no.~10, pp. 2024--2039, 2015.

\bibitem{garg2016unsupervised}
R.~Garg, V.~K. BG, G.~Carneiro, and I.~Reid, ``Unsupervised {CNN} for single
  view depth estimation: Geometry to the rescue,'' in \emph{ECCV}, 2016, pp.
  740--756.

\bibitem{godard2017unsupervised}
C.~Godard, O.~Mac~Aodha, and G.~J. Brostow, ``Unsupervised monocular depth
  estimation with left-right consistency,'' in \emph{CVPR}, 2017, pp. 270--279.

\bibitem{zhou2017unsupervised}
T.~Zhou, M.~Brown, N.~Snavely, and D.~G. Lowe, ``Unsupervised learning of depth
  and ego-motion from video,'' in \emph{CVPR}, 2017, pp. 1851--1858.

\bibitem{kuznietsov2017semi}
Y.~Kuznietsov, J.~Stuckler, and B.~Leibe, ``Semi-supervised deep learning for
  monocular depth map prediction,'' in \emph{CVPR}, 2017, pp. 6647--6655.

\bibitem{aleotti2018generative}
F.~Aleotti, F.~Tosi, M.~Poggi, and S.~Mattoccia, ``Generative adversarial
  networks for unsupervised monocular depth prediction,'' in \emph{ECCV
  Workshops}, 2018, pp. 337--354.

\bibitem{poggi2018towards}
M.~Poggi, F.~Aleotti, F.~Tosi, and S.~Mattoccia, ``Towards real-time
  unsupervised monocular depth estimation on {CPU},'' in \emph{IROS}, 2018, pp.
  5848--5854.

\bibitem{spek2018cream}
A.~Spek, T.~Dharmasiri, and T.~Drummond, ``{CReaM}: Condensed real-time models
  for depth prediction using convolutional neural networks,'' in \emph{IROS},
  2018, pp. 540--547.

\bibitem{bhat2021adabins}
S.~F. Bhat, I.~Alhashim, and P.~Wonka, ``Adabins: depth estimation using
  adaptive bins,'' in \emph{CVPR}, 2021, pp. 4009--4018.

\bibitem{ullman1979interpretation}
S.~Ullman, ``The interpretation of structure from motion,'' \emph{Proceedings
  of the Royal Society of London. Series B. Biological Sciences}, vol. 203, no.
  1153, pp. 405--426, 1979.

\bibitem{horn1970shape}
B.~K. Horn, ``Shape from shading: A method for obtaining the shape of a smooth
  opaque object from one view,'' 1970.

\bibitem{barrow1978recovering}
H.~Barrow, J.~Tenenbaum, A.~Hanson, and E.~Riseman, ``Recovering intrinsic
  scene characteristics,'' \emph{Comput. Vis. Syst}, vol.~2, no. 3-26, p.~2,
  1978.

\bibitem{kong2015intrinsic}
N.~Kong and M.~J. Black, ``Intrinsic depth: Improving depth transfer with
  intrinsic images,'' in \emph{ICCV}, 2015, pp. 3514--3522.

\bibitem{ma2018sparse}
F.~Ma and S.~Karaman, ``Sparse-to-dense: Depth prediction from sparse depth
  samples and a single image,'' in \emph{ICRA}, 2018, pp. 4796--4803.

\bibitem{cheng2018depth}
X.~Cheng, P.~Wang, and R.~Yang, ``Depth estimation via affinity learned with
  convolutional spatial propagation network,'' in \emph{ECCV}, 2018, pp.
  103--119.

\bibitem{chen2019structure}
X.~Chen, X.~Chen, and Z.-J. Zha, ``Structure-aware residual pyramid network for
  monocular depth estimation,'' \emph{arXiv preprint arXiv:1907.06023}, 2019.

\bibitem{li2015depth}
B.~Li, C.~Shen, Y.~Dai, A.~Van Den~Hengel, and M.~He, ``Depth and surface
  normal estimation from monocular images using regression on deep features and
  hierarchical {CRFs},'' in \emph{CVPR}, 2015, pp. 1119--1127.

\bibitem{heo2018monocular}
M.~Heo, J.~Lee, K.-R. Kim, H.-U. Kim, and C.-S. Kim, ``Monocular depth
  estimation using whole strip masking and reliability-based refinement,'' in
  \emph{ECCV}, 2018, pp. 36--51.

\bibitem{mehta2018structured}
I.~Mehta, P.~Sakurikar, and P.~Narayanan, ``Structured adversarial training for
  unsupervised monocular depth estimation,'' in \emph{3DV}, 2018, pp. 314--323.

\bibitem{he2016deep}
K.~He, X.~Zhang, S.~Ren, and J.~Sun, ``Deep residual learning for image
  recognition,'' in \emph{CVPR}, 2016, pp. 770--778.

\bibitem{huang2017densely}
G.~Huang, Z.~Liu, L.~Van Der~Maaten, and K.~Q. Weinberger, ``Densely connected
  convolutional networks,'' in \emph{CVPR}, 2017, pp. 4700--4708.

\bibitem{hu2018squeeze}
J.~Hu, L.~Shen, and G.~Sun, ``Squeeze-and-excitation networks,'' in
  \emph{CVPR}, 2018, pp. 7132--7141.

\bibitem{mancini2016fast}
M.~Mancini, G.~Costante, P.~Valigi, and T.~A. Ciarfuglia, ``Fast robust
  monocular depth estimation for obstacle detection with fully convolutional
  networks,'' in \emph{IROS}, 2016, pp. 4296--4303.

\bibitem{alhashim2018high}
I.~Alhashim and P.~Wonka, ``High quality monocular depth estimation via
  transfer learning,'' \emph{arXiv preprint arXiv:1812.11941}, 2018.

\bibitem{lee2019big}
J.~H. Lee, M.-K. Han, D.~W. Ko, and I.~H. Suh, ``From big to small: Multi-scale
  local planar guidance for monocular depth estimation,'' \emph{arXiv preprint
  arXiv:1907.10326}, 2019.

\bibitem{yin2019enforcing}
W.~Yin, Y.~Liu, C.~Shen, and Y.~Yan, ``Enforcing geometric constraints of
  virtual normal for depth prediction,'' in \emph{ICCV}, 2019, pp. 5684--5693.

\bibitem{fu2018deep}
H.~Fu, M.~Gong, C.~Wang, K.~Batmanghelich, and D.~Tao, ``Deep ordinal
  regression network for monocular depth estimation,'' in \emph{CVPR}, 2018,
  pp. 2002--2011.

\bibitem{vaswani2017attention}
A.~Vaswani, N.~Shazeer, N.~Parmar, J.~Uszkoreit, L.~Jones, A.~N. Gomez,
  L.~Kaiser, and I.~Polosukhin, ``Attention is all you need,'' \emph{arXiv
  preprint arXiv:1706.03762}, 2017.

\bibitem{dosovitskiy2020image}
A.~Dosovitskiy, L.~Beyer, A.~Kolesnikov, D.~Weissenborn, X.~Zhai,
  T.~Unterthiner, M.~Dehghani, M.~Minderer, G.~Heigold, S.~Gelly \emph{et~al.},
  ``An image is worth 16x16 words: Transformers for image recognition at
  scale,'' \emph{arXiv preprint arXiv:2010.11929}, 2020.

\bibitem{wang2020bifuse}
F.-E. Wang, Y.-H. Yeh, M.~Sun, W.-C. Chiu, and Y.-H. Tsai, ``Bifuse: Monocular
  360 depth estimation via bi-projection fusion,'' in \emph{CVPR}, 2020, pp.
  462--471.

\bibitem{cs2018depthnet}
A.~CS~Kumar, S.~M. Bhandarkar, and M.~Prasad, ``{DepthNet}: A recurrent neural
  network architecture for monocular depth prediction,'' in \emph{CVPR
  Workshops}, 2018, pp. 283--291.

\bibitem{zhang2019exploiting}
H.~Zhang, C.~Shen, Y.~Li, Y.~Cao, Y.~Liu, and Y.~Yan, ``Exploiting temporal
  consistency for real-time video depth estimation,'' in \emph{ICCV}, 2019, pp.
  1725--1734.

\bibitem{wang2019recurrent}
R.~Wang, S.~M. Pizer, and J.-M. Frahm, ``Recurrent neural network for {(Un-)}
  supervised learning of monocular video visual odometry and depth,'' in
  \emph{CVPR}, 2019, pp. 5555--5564.

\bibitem{li2018deep}
R.~Li, K.~Xian, C.~Shen, Z.~Cao, H.~Lu, and L.~Hang, ``Deep attention-based
  classification network for robust depth prediction,'' in \emph{ACCV}, 2018,
  pp. 663--678.

\bibitem{chen2019attention}
Y.~Chen, H.~Zhao, and Z.~Hu, ``Attention-based context aggregation network for
  monocular depth estimation,'' \emph{arXiv preprint arXiv:1901.10137}, 2019.

\bibitem{zou2019mean}
H.~Zou, K.~Xian, J.~Yang, and Z.~Cao, ``Mean-variance loss for monocular depth
  estimation,'' in \emph{ICIP}, 2019, pp. 1760--1764.

\bibitem{liebel2019multidepth}
L.~Liebel and M.~K{\"o}rner, ``{MultiDepth}: Single-image depth estimation via
  multi-task regression and classification,'' \emph{arXiv preprint
  arXiv:1907.11111}, 2019.

\bibitem{song2019contextualized}
W.~Song, S.~Li, J.~Liu, A.~Hao, Q.~Zhao, and H.~Qin, ``Contextualized {CNN} for
  scene-aware depth estimation from single {RGB} image,'' \emph{IEEE TMM},
  vol.~22, no.~5, pp. 1220--1233, 2019.

\bibitem{sanchez2018hybridnet}
D.~S{\'a}nchez-Escobedo, X.~Lin, J.~R. Casas, and M.~Pardas, ``{HybridNet} for
  depth estimation and semantic segmentation,'' in \emph{ICASSP}, 2018, pp.
  1563--1567.

\bibitem{wang2015towards}
P.~Wang, X.~Shen, Z.~Lin, S.~Cohen, B.~Price, and A.~L. Yuille, ``Towards
  unified depth and semantic prediction from a single image,'' in \emph{CVPR},
  2015, pp. 2800--2809.

\bibitem{jafari2017analyzing}
O.~H. Jafari, O.~Groth, A.~Kirillov, M.~Y. Yang, and C.~Rother, ``Analyzing
  modular {CNN} architectures for joint depth prediction and semantic
  segmentation,'' in \emph{ICRA}, 2017, pp. 4620--4627.

\bibitem{gurram2018monocular}
A.~Gurram, O.~Urfalioglu, I.~Halfaoui, F.~Bouzaraa, and A.~M. L{\'o}pez,
  ``Monocular depth estimation by learning from heterogeneous datasets,'' in
  \emph{IV}, 2018, pp. 2176--2181.

\bibitem{jiao2018look}
J.~Jiao, Y.~Cao, Y.~Song, and R.~Lau, ``Look deeper into depth: Monocular depth
  estimation with semantic booster and attention-driven loss,'' in \emph{ECCV},
  2018, pp. 53--69.

\bibitem{qi2018geonet}
X.~Qi, R.~Liao, Z.~Liu, R.~Urtasun, and J.~Jia, ``{GeoNet}: Geometric neural
  network for joint depth and surface normal estimation,'' in \emph{CVPR},
  2018, pp. 283--291.

\bibitem{nekrasov2019real}
V.~Nekrasov, T.~Dharmasiri, A.~Spek, T.~Drummond, C.~Shen, and I.~Reid,
  ``Real-time joint semantic segmentation and depth estimation using asymmetric
  annotations,'' in \emph{ICRA}, 2019, pp. 7101--7107.

\bibitem{hsieh2019deep}
Y.-Y. Hsieh, W.-Y. Lin, D.-L. Li, and J.-H. Chuang, ``Deep learning-based
  obstacle detection and depth estimation,'' in \emph{ICIP}, 2019, pp.
  1635--1639.

\bibitem{abdulwahab2020adversarial}
S.~Abdulwahab, H.~A. Rashwan, M.~A. Garcia, M.~Jabreel, S.~Chambon, and
  D.~Puig, ``Adversarial learning for depth and viewpoint estimation from a
  single image,'' \emph{IEEE TCSVT}, 2020.

\bibitem{long2015fully}
J.~Long, E.~Shelhamer, and T.~Darrell, ``Fully convolutional networks for
  semantic segmentation,'' in \emph{CVPR}, 2015, pp. 3431--3440.

\bibitem{redmon2018yolov3}
J.~Redmon and A.~Farhadi, ``{YoloV3}: An incremental improvement,'' \emph{arXiv
  preprint arXiv:1804.02767}, 2018.

\bibitem{romera2017erfnet}
E.~Romera, J.~M. Alvarez, L.~M. Bergasa, and R.~Arroyo, ``{ERFNet}: Efficient
  residual factorized convnet for real-time semantic segmentation,'' \emph{IEEE
  TITS}, vol.~19, no.~1, pp. 263--272, 2017.

\bibitem{wang2020depthnet}
L.~Wang, M.~Famouri, and A.~Wong, ``{DepthNet Nano}: A highly compact
  self-normalizing neural network for monocular depth estimation,'' \emph{arXiv
  preprint arXiv:2004.08008}, 2020.

\bibitem{mahjourian2018unsupervised}
R.~Mahjourian, M.~Wicke, and A.~Angelova, ``Unsupervised learning of depth and
  ego-motion from monocular video using {3D} geometric constraints,'' in
  \emph{CVPR}, 2018, pp. 5667--5675.

\bibitem{tosi2019learning}
F.~Tosi, F.~Aleotti, M.~Poggi, and S.~Mattoccia, ``Learning monocular depth
  estimation infusing traditional stereo knowledge,'' in \emph{CVPR}, 2019, pp.
  9799--9809.

\bibitem{ma2019self}
F.~Ma, G.~V. Cavalheiro, and S.~Karaman, ``Self-supervised sparse-to-dense:
  Self-supervised depth completion from {LiDAR} and monocular camera,'' in
  \emph{ICRA}, 2019, pp. 3288--3295.

\bibitem{zhang2019dfinenet}
Y.~Zhang, T.~Nguyen, I.~D. Miller, S.~Chen, C.~J. Taylor, V.~Kumar
  \emph{et~al.}, ``{DFineNet}: Ego-motion estimation and depth refinement from
  sparse, noisy depth input with {RGB} guidance,'' \emph{arXiv preprint
  arXiv:1903.06397}, 2019.

\bibitem{fei2019geo}
X.~Fei, A.~Wong, and S.~Soatto, ``Geo-supervised visual depth prediction,''
  \emph{IEEE RAL}, vol.~4, no.~2, pp. 1661--1668, 2019.

\bibitem{guizilini2020semantically}
V.~Guizilini, R.~Hou, J.~Li, R.~Ambrus, and A.~Gaidon, ``Semantically-guided
  representation learning for self-supervised monocular depth,'' \emph{arXiv
  preprint arXiv:2002.12319}, 2020.

\bibitem{guizilini20203d}
V.~Guizilini, R.~Ambrus, S.~Pillai, A.~Raventos, and A.~Gaidon, ``{3D} packing
  for self-supervised monocular depth estimation,'' in \emph{CVPR}, 2020, pp.
  2485--2494.

\bibitem{johnston2020self}
A.~Johnston and G.~Carneiro, ``Self-supervised monocular trained depth
  estimation using self-attention and discrete disparity volume,'' in
  \emph{CVPR}, 2020, pp. 4756--4765.

\bibitem{mayer2016large}
N.~Mayer, E.~Ilg, P.~Hausser, P.~Fischer, D.~Cremers, A.~Dosovitskiy, and
  T.~Brox, ``A large dataset to train convolutional networks for disparity,
  optical flow, and scene flow estimation,'' in \emph{CVPR}, 2016, pp.
  4040--4048.

\bibitem{prasad2019sfmlearner++}
V.~Prasad and B.~Bhowmick, ``{SfMLearner++}: Learning monocular depth \&
  ego-motion using meaningful geometric constraints,'' in \emph{WACV}, 2019,
  pp. 2087--2096.

\bibitem{klodt2018supervising}
M.~Klodt and A.~Vedaldi, ``Supervising the new with the old: learning {SfM}
  from {SfM},'' in \emph{ECCV}, 2018, pp. 698--713.

\bibitem{mur2017orb}
R.~Mur-Artal and J.~D. Tard{\'o}s, ``{ORB-SLAM2}: An open-source {SLAM} system
  for monocular, stereo, and {RGB-D} cameras,'' \emph{IEEE TRO}, vol.~33,
  no.~5, pp. 1255--1262, 2017.

\bibitem{vijayanarasimhan2017sfm}
S.~Vijayanarasimhan, S.~Ricco, C.~Schmid, R.~Sukthankar, and K.~Fragkiadaki,
  ``{SfM-Net}: Learning of structure and motion from video,'' \emph{arXiv
  preprint arXiv:1704.07804}, 2017.

\bibitem{dai2019self}
Q.~Dai, V.~Patil, S.~Hecker, D.~Dai, L.~Van~Gool, and K.~Schindler,
  ``Self-supervised object motion and depth estimation from video,''
  \emph{arXiv preprint arXiv:1912.04250}, 2019.

\bibitem{bian2019unsupervised}
J.-W. Bian, Z.~Li, N.~Wang, H.~Zhan, C.~Shen, M.-M. Cheng, and I.~Reid,
  ``Unsupervised scale-consistent depth and ego-motion learning from monocular
  video,'' \emph{arXiv preprint arXiv:1908.10553}, 2019.

\bibitem{zhao2020towards}
W.~Zhao, S.~Liu, Y.~Shu, and Y.-J. Liu, ``Towards better generalization: joint
  depth-pose learning without posenet,'' \emph{arXiv preprint
  arXiv:2004.01314}, 2020.

\bibitem{kendall2015posenet}
A.~Kendall, M.~Grimes, and R.~Cipolla, ``{PoseNet}: A convolutional network for
  real-time 6-{DoF} camera relocalization,'' in \emph{ICCV}, 2015, pp.
  2938--2946.

\bibitem{zou2018df}
Y.~Zou, Z.~Luo, and J.-B. Huang, ``{DF-Net}: Unsupervised joint learning of
  depth and flow using cross-task consistency,'' in \emph{ECCV}, 2018, pp.
  36--53.

\bibitem{yin2018geonet}
Z.~Yin and J.~Shi, ``{GeoNet}: Unsupervised learning of dense depth, optical
  flow and camera pose,'' in \emph{CVPR}, 2018, pp. 1983--1992.

\bibitem{ranjan2019competitive}
A.~Ranjan, V.~Jampani, L.~Balles, K.~Kim, D.~Sun, J.~Wulff, and M.~J. Black,
  ``Competitive collaboration: Joint unsupervised learning of depth, camera
  motion, optical flow and motion segmentation,'' in \emph{CVPR}, 2019, pp.
  12\,240--12\,249.

\bibitem{pilzer2018unsupervised}
A.~Pilzer, D.~Xu, M.~Puscas, E.~Ricci, and N.~Sebe, ``Unsupervised adversarial
  depth estimation using cycled generative networks,'' in \emph{3DV}, 2018, pp.
  587--595.

\bibitem{wang2020adversarial}
A.~Wang, Z.~Fang, Y.~Gao, S.~Tan, S.~Wang, S.~Ma, and J.-N. Hwang,
  ``Adversarial learning for joint optimization of depth and ego-motion,''
  \emph{IEEE TIP}, vol.~29, pp. 4130--4142, 2020.

\bibitem{almalioglu2019ganvo}
Y.~Almalioglu, M.~R.~U. Saputra, P.~P. de~Gusmao, A.~Markham, and N.~Trigoni,
  ``{GANVO}: Unsupervised deep monocular visual odometry and depth estimation
  with generative adversarial networks,'' in \emph{ICRA}, 2019, pp. 5474--5480.

\bibitem{liu2020mininet}
J.~Liu, Q.~Li, R.~Cao, W.~Tang, and G.~Qiu, ``{MiniNet}: An extremely
  lightweight convolutional neural network for real-time unsupervised monocular
  depth estimation,'' \emph{ISPRS JPRS}, vol. 166, pp. 255--267, 2020.

\bibitem{guo2018learning}
X.~Guo, H.~Li, S.~Yi, J.~Ren, and X.~Wang, ``Learning monocular depth by
  distilling cross-domain stereo networks,'' in \emph{ECCV}, 2018, pp.
  484--500.

\bibitem{qiu2019deeplidar}
J.~Qiu, Z.~Cui, Y.~Zhang, X.~Zhang, S.~Liu, B.~Zeng, and M.~Pollefeys,
  ``{DeepLiDAR}: Deep surface normal guided depth prediction for outdoor scene
  from sparse {LiDAR} data and single color image,'' in \emph{CVPR}, 2019, pp.
  3313--3322.

\bibitem{fang2020towards}
Z.~Fang, X.~Chen, Y.~Chen, and L.~V. Gool, ``Towards good practice for
  {CNN}-based monocular depth estimation,'' in \emph{WACV}, 2020, pp.
  1091--1100.

\bibitem{sartipi2020deep}
K.~Sartipi, T.~Do, T.~Ke, K.~Vuong, and S.~I. Roumeliotis, ``Deep depth
  estimation from visual-inertial {SLAM},'' \emph{arXiv preprint
  arXiv:2008.00092}, 2020.

\bibitem{xian2020structure}
K.~Xian, J.~Zhang, O.~Wang, L.~Mai, Z.~Lin, and Z.~Cao, ``Structure-guided
  ranking loss for single image depth prediction,'' in \emph{CVPR}, 2020, pp.
  611--620.

\bibitem{zhan2018unsupervised}
H.~Zhan, R.~Garg, C.~Saroj~Weerasekera, K.~Li, H.~Agarwal, and I.~Reid,
  ``Unsupervised learning of monocular depth estimation and visual odometry
  with deep feature reconstruction,'' in \emph{CVPR}, 2018, pp. 340--349.

\bibitem{casser2019depth}
V.~Casser, S.~Pirk, R.~Mahjourian, and A.~Angelova, ``Depth prediction without
  the sensors: Leveraging structure for unsupervised learning from monocular
  videos,'' in \emph{AAAI}, vol.~33, 2019, pp. 8001--8008.

\bibitem{elkerdawy2019lightweight}
S.~Elkerdawy, H.~Zhang, and N.~Ray, ``Lightweight monocular depth estimation
  model by joint end-to-end filter pruning,'' \emph{arXiv preprint
  arXiv:1905.05212}, 2019.

\bibitem{godard2019digging}
C.~Godard, O.~Mac~Aodha, M.~Firman, and G.~J. Brostow, ``Digging into
  self-supervised monocular depth estimation,'' in \emph{ICCV}, 2019, pp.
  3828--3838.

\bibitem{watson2019self}
J.~Watson, M.~Firman, G.~J. Brostow, and D.~Turmukhambetov, ``Self-supervised
  monocular depth hints,'' in \emph{ICCV}, 2019, pp. 2162--2171.

\bibitem{zioulis2019spherical}
N.~Zioulis, A.~Karakottas, D.~Zarpalas, F.~Alvarez, and P.~Daras, ``Spherical
  view synthesis for self-supervised 360° depth estimation,'' in \emph{3DV},
  2019, pp. 690--699.

\bibitem{klingner2020self}
M.~Klingner, J.-A. Term{\"o}hlen, J.~Mikolajczyk, and T.~Fingscheidt,
  ``Self-supervised monocular depth estimation: Solving the dynamic object
  problem by semantic guidance,'' \emph{arXiv preprint arXiv:2007.06936}, 2020.

\bibitem{peng2019edge}
K.-S. Peng, G.~Ditzler, and J.~Rozenblit, ``Edge-guided occlusion fading
  reduction for a light-weighted self-supervised monocular depth estimation,''
  \emph{arXiv preprint arXiv:1911.11705}, 2019.

\bibitem{shu2020feature}
C.~Shu, K.~Yu, Z.~Duan, and K.~Yang, ``Feature-metric loss for self-supervised
  learning of depth and egomotion,'' \emph{arXiv preprint arXiv:2007.10603},
  2020.

\bibitem{xue2020toward}
F.~Xue, G.~Zhuo, Z.~Huang, W.~Fu, Z.~Wu, and M.~H. Ang~Jr, ``Toward
  hierarchical self-supervised monocular absolute depth estimation for
  autonomous driving applications,'' \emph{arXiv preprint arXiv:2004.05560},
  2020.

\bibitem{ramirez2018geometry}
P.~Z. Ramirez, M.~Poggi, F.~Tosi, S.~Mattoccia, and L.~Di~Stefano, ``Geometry
  meets semantics for semi-supervised monocular depth estimation,'' in
  \emph{ACCV}, 2018, pp. 298--313.

\bibitem{amiri2019semi}
A.~J. Amiri, S.~Y. Loo, and H.~Zhang, ``Semi-supervised monocular depth
  estimation with left-right consistency using deep neural network,'' in
  \emph{ROBIO}, 2019, pp. 602--607.

\bibitem{atapour2018real}
A.~Atapour-Abarghouei and T.~P. Breckon, ``Real-time monocular depth estimation
  using synthetic data with domain adaptation via image style transfer,'' in
  \emph{CVPR}, 2018, pp. 2800--2810.

\bibitem{zheng2018t2net}
C.~Zheng, T.-J. Cham, and J.~Cai, ``{T2Net}: Synthetic-to-realistic translation
  for solving single-image depth estimation tasks,'' in \emph{ECCV}, 2018, pp.
  767--783.

\bibitem{zhao2019geometry}
S.~Zhao, H.~Fu, M.~Gong, and D.~Tao, ``Geometry-aware symmetric domain
  adaptation for monocular depth estimation,'' in \emph{CVPR}, 2019, pp.
  9788--9798.

\bibitem{patil2020don}
V.~Patil, W.~Van~Gansbeke, D.~Dai, and L.~Van~Gool, ``Don’t forget the past:
  Recurrent depth estimation from monocular video,'' \emph{IEEE RAL}, vol.~5,
  no.~4, pp. 6813--6820, 2020.

\bibitem{dos2019sparse}
N.~dos Santos~Rosa, V.~Guizilini, and V.~Grassi, ``Sparse-to-continuous:
  Enhancing monocular depth estimation using occupancy maps,'' in \emph{2019
  19th International Conference on Advanced Robotics (ICAR)}, 2019, pp.
  793--800.

\bibitem{guizilini2020robust}
V.~Guizilini, J.~Li, R.~Ambrus, S.~Pillai, and A.~Gaidon, ``Robust
  semi-supervised monocular depth estimation with reprojected distances,'' in
  \emph{CoRL}, 2020, pp. 503--512.

\bibitem{zhao2020domain}
Y.~Zhao, S.~Kong, D.~Shin, and C.~Fowlkes, ``Domain decluttering: simplifying
  images to mitigate synthetic-real domain shift and improve depth
  estimation,'' in \emph{CVPR}, 2020, pp. 3330--3340.

\bibitem{tian2019semi}
H.~Tian and F.~Li, ``Semi-supervised depth estimation from a single image based
  on confidence learning,'' in \emph{ICASSP}, 2019, pp. 8573--8577.

\bibitem{ji2019semi}
R.~Ji, K.~Li, Y.~Wang, X.~Sun, F.~Guo, X.~Guo, Y.~Wu, F.~Huang, and J.~Luo,
  ``Semi-supervised adversarial monocular depth estimation,'' \emph{IEEE
  TPAMI}, vol.~42, no.~10, pp. 2410--2422, 2019.

\bibitem{yuesemi}
M.~Yue, G.~Fu, M.~Wu, and H.~Wang, ``Semi-supervised monocular depth estimation
  based on semantic supervision,'' \emph{JIRS}, 2020.

\bibitem{pang2017cascade}
J.~Pang, W.~Sun, J.~S. Ren, C.~Yang, and Q.~Yan, ``Cascade residual learning: A
  two-stage convolutional neural network for stereo matching,'' in \emph{ICCV},
  2017, pp. 887--895.

\bibitem{tonioni2019unsupervised}
A.~Tonioni, M.~Poggi, S.~Mattoccia, and L.~Di~Stefano, ``Unsupervised domain
  adaptation for depth prediction from images,'' \emph{IEEE TPAMI}, 2019.

\bibitem{jing2019neural}
Y.~Jing, Y.~Yang, Z.~Feng, J.~Ye, Y.~Yu, and M.~Song, ``Neural style transfer:
  A review,'' \emph{IEEE TVCG}, vol.~26, no.~11, pp. 3365--3385, 2019.

\bibitem{liao2017parse}
Y.~Liao, L.~Huang, Y.~Wang, S.~Kodagoda, Y.~Yu, and Y.~Liu, ``Parse geometry
  from a line: Monocular depth estimation with partial laser observation,'' in
  \emph{ICRA}, 2017, pp. 5059--5066.

\bibitem{jaritz2018sparse}
M.~Jaritz, R.~De~Charette, E.~Wirbel, X.~Perrotton, and F.~Nashashibi, ``Sparse
  and dense data with {CNNs}: Depth completion and semantic segmentation,'' in
  \emph{3DV}, 2018, pp. 52--60.

\bibitem{wang2019plug}
T.-H. Wang, F.-E. Wang, J.-T. Lin, Y.-H. Tsai, W.-C. Chiu, and M.~Sun,
  ``Plug-and-play: Improve depth prediction via sparse data propagation,'' in
  \emph{ICRA}, 2019, pp. 5880--5886.

\bibitem{chen2019learning}
Y.~Chen, B.~Yang, M.~Liang, and R.~Urtasun, ``Learning joint {2D-3D}
  representations for depth completion,'' in \emph{ICCV}, 2019, pp.
  10\,023--10\,032.

\bibitem{yurtsever2020survey}
E.~Yurtsever, J.~Lambert, A.~Carballo, and K.~Takeda, ``A survey of autonomous
  driving: Common practices and emerging technologies,'' \emph{IEEE Access},
  vol.~8, pp. 58\,443--58\,469, 2020.

\bibitem{howard2017mobilenets}
A.~G. Howard, M.~Zhu, B.~Chen, D.~Kalenichenko, W.~Wang, T.~Weyand,
  M.~Andreetto, and H.~Adam, ``{MobileNets}: Efficient convolutional neural
  networks for mobile vision applications,'' \emph{arXiv preprint
  arXiv:1704.04861}, 2017.

\bibitem{sandler2018mobilenetv2}
M.~Sandler, A.~Howard, M.~Zhu, A.~Zhmoginov, and L.-C. Chen, ``{MobileNetV2}:
  Inverted residuals and linear bottlenecks,'' in \emph{CVPR}, 2018, pp.
  4510--4520.

\bibitem{oh2020rrnet}
S.~Oh, H.-J.~S. Kim, J.~Lee, and J.~Kim, ``{RRNet}: Repetition-reduction
  network for energy efficient depth estimation,'' \emph{IEEE Access}, vol.~8,
  pp. 106\,097--106\,108, 2020.

\bibitem{yin2017scale}
X.~Yin, X.~Wang, X.~Du, and Q.~Chen, ``Scale recovery for monocular visual
  odometry using depth estimated with deep convolutional neural fields,'' in
  \emph{ICCV}, 2017, pp. 5870--5878.

\bibitem{luo2018real}
H.~Luo, Y.~Gao, Y.~Wu, C.~Liao, X.~Yang, and K.-T. Cheng, ``Real-time dense
  monocular {SLAM} with online adapted depth prediction network,'' \emph{IEEE
  TMM}, vol.~21, no.~2, pp. 470--483, 2018.

\bibitem{yang2018deep}
N.~Yang, R.~Wang, J.~Stuckler, and D.~Cremers, ``Deep virtual stereo odometry:
  Leveraging deep depth prediction for monocular direct sparse odometry,'' in
  \emph{ECCV}, 2018, pp. 817--833.

\bibitem{mendes2020deep}
R.~d.~Q. Mendes, E.~G. Ribeiro, N.~d.~S. Rosa, and V.~Grassi~Jr, ``On deep
  learning techniques to boost monocular depth estimation for autonomous
  navigation,'' \emph{arXiv preprint arXiv:2010.06626}, 2020.

\bibitem{michels2005high}
J.~Michels, A.~Saxena, and A.~Y. Ng, ``High speed obstacle avoidance using
  monocular vision and reinforcement learning,'' in \emph{ICML}, 2005, pp.
  593--600.

\bibitem{alvarez2016collision}
H.~Alvarez, L.~M. Paz, J.~Sturm, and D.~Cremers, ``Collision avoidance for
  quadrotors with a monocular camera,'' in \emph{Experimental Robotics}, 2016,
  pp. 195--209.

\bibitem{chakravarty2017cnn}
P.~Chakravarty, K.~Kelchtermans, T.~Roussel, S.~Wellens, T.~Tuytelaars, and
  L.~Van~Eycken, ``{CNN}-based single image obstacle avoidance on a
  quadrotor,'' in \emph{ICRA}, 2017, pp. 6369--6374.

\bibitem{zhang2019monocular}
Z.~Zhang, M.~Xiong, and H.~Xiong, ``Monocular depth estimation for {UAV}
  obstacle avoidance,'' in \emph{CCIOT}, 2019, pp. 43--47.

\bibitem{lin2020robust}
J.~Lin, H.~Zhu, and J.~Alonso-Mora, ``Robust vision-based obstacle avoidance
  for micro aerial vehicles in dynamic environments,'' \emph{arXiv preprint
  arXiv:2002.04920}, 2020.

\bibitem{scharwachter2015low}
T.~Scharw{\"a}chter and U.~Franke, ``Low-level fusion of color, texture and
  depth for robust road scene understanding,'' in \emph{IV}, 2015, pp.
  599--604.

\bibitem{rojas2018landingzone}
L.~RoJas-Perez, R.~Munguia-Silva, and J.~Martinez-Carranza, ``Real-time landing
  zone detection for {UAVs} using single aerial images,'' in \emph{IMAVCC},
  2018, pp. 243--248.

\bibitem{li2016metric}
P.~Li, M.~Garratt, A.~Lambert, and S.~Lin, ``Metric sensing and control of a
  quadrotor using a homography-based visual inertial fusion method,''
  \emph{RAS}, vol.~76, pp. 1--14, 2016.

\bibitem{wang2015real}
J.~Wang, M.~Garratt, S.~Anavatti, and S.~Lin, ``Real-time visual odometry for
  autonomous mav navigation using rgb-d camera,'' in \emph{ROBIO}, 2015, pp.
  1353--1358.

\bibitem{engel2014lsd}
J.~Engel, T.~Sch{\"o}ps, and D.~Cremers, ``{LSD-SLAM}: Large-scale direct
  monocular {SLAM},'' in \emph{ECCV}, 2014, pp. 834--849.

\bibitem{loo2019cnn}
S.~Y. Loo, A.~J. Amiri, S.~Mashohor, S.~H. Tang, and H.~Zhang, ``{CNN-SVO}:
  Improving the mapping in semi-direct visual odometry using single-image depth
  prediction,'' in \emph{ICRA}, 2019, pp. 5218--5223.

\bibitem{matthies2014stereo}
L.~Matthies, R.~Brockers, Y.~Kuwata, and S.~Weiss, ``Stereo vision-based
  obstacle avoidance for micro air vehicles using disparity space,'' in
  \emph{ICRA}, 2014, pp. 3242--3249.

\bibitem{brockers2016vision}
R.~Brockers, A.~Fragoso, B.~Rothrock, C.~Lee, and L.~Matthies, ``Vision-based
  obstacle avoidance for micro air vehicles using an egocylindrical depth
  map,'' in \emph{ISER}, 2016, pp. 505--514.

\bibitem{wojke2012moving}
N.~Wojke and M.~H{\"a}selich, ``Moving vehicle detection and tracking in
  unstructured environments,'' in \emph{ICRA}, 2012, pp. 3082--3087.

\bibitem{wang2012could}
D.~Z. Wang, I.~Posner, and P.~Newman, ``What could move? finding cars,
  pedestrians and bicyclists in 3d laser data,'' in \emph{ICRA}, 2012, pp.
  4038--4044.

\bibitem{cherubini2014autonomous}
A.~Cherubini, F.~Spindler, and F.~Chaumette, ``Autonomous visual navigation and
  laser-based moving obstacle avoidance,'' \emph{IEEE TITS}, vol.~15, no.~5,
  pp. 2101--2110, 2014.

\bibitem{hu2019visualization}
J.~Hu, Y.~Zhang, and T.~Okatani, ``Visualization of convolutional neural
  networks for monocular depth estimation,'' in \emph{ICCV}, 2019, pp.
  3869--3878.

\bibitem{dijk2019neural}
T.~v. Dijk and G.~d. Croon, ``How do neural networks see depth in single
  images?'' in \emph{ICCV}, 2019, pp. 2183--2191.

\end{thebibliography}
\bibliographystyle{IEEEtran}

\end{document}